\documentclass[letterpaper]{article} 
\usepackage{aaai25}  

\usepackage{times}  
\usepackage{helvet}  
\usepackage{courier}  
\usepackage[hyphens]{url}  
\usepackage{graphicx} 
\usepackage{natbib}  
\usepackage{caption} 
\usepackage{algorithm}
\usepackage{algorithmic}
\usepackage{makecell}
\usepackage{newfloat}
\usepackage{listings}
\DeclareCaptionStyle{ruled}{labelfont=normalfont,labelsep=colon,strut=off} 
\floatstyle{ruled}
\newfloat{listing}{tb}{lst}{}
\floatname{listing}{Listing}
\usepackage[svgnames,x11names,table]{xcolor}
\usepackage[pagebackref,breaklinks,colorlinks,citecolor=DodgerBlue4]{hyperref} 

\usepackage{subcaption}
\usepackage{array}
\usepackage{adjustbox}
\usepackage{colortbl}
\usepackage{booktabs}
\usepackage{multirow}
\usepackage{arydshln}
\usepackage{subcaption}

\newcommand{\qt}[1]{\textcolor{black}{#1}} 
\newcommand{\kt}[1]{\textcolor{black}{#1}} 

\usepackage{pbox}

\nocopyright
\title{TB-Bench: Training and Testing Multi-Modal AI for Understanding Spatio-Temporal Traffic Behaviors from Dashcam Images/Videos}
\author {
    Korawat Charoenpitaks\textsuperscript{\rm 1,*},
    Van-Quang Nguyen\textsuperscript{\rm 2,*},
    Masanori Suganuma\textsuperscript{\rm 1},
    Kentaro Arai\textsuperscript{\rm 3},
    Seiji Totsuka\textsuperscript{\rm 3},
    Hiroshi Ino\textsuperscript{\rm 3},
    Takayuki Okatani\textsuperscript{\rm 1,2}
}
\affiliations {
    \textsuperscript{\rm 1}Tohoku University\\
    \textsuperscript{\rm 2}RIKEN Center for AIP\\
    \textsuperscript{\rm 3}DENSO CORPORATION\\
    korawat@vision.is.tohoku.ac.jp, 
    quang.nguyen.jz@riken.jp, 
    suganuma@vision.is.tohoku.ac.jp, 
    kentaro.arai.j8n@jp.denso.com,
    seiji.totsuka.j7z@jp.denso.com,
    hiroshi.naganawa.j7v@jp.denso.com,
    okatani@tohoku.ac.jp
    \vspace{0.2cm}
    \textsuperscript{*}Corresponding authors
}

\usepackage{bibentry}

\begin{document}

\maketitle
\begin{abstract}
The application of Multi-modal Large Language Models (MLLMs) in Autonomous Driving (AD) faces significant challenges due to their limited training on traffic-specific data and the absence of dedicated benchmarks for spatiotemporal understanding. This study addresses these issues by proposing TB-Bench, a comprehensive benchmark designed to evaluate MLLMs on understanding traffic behaviors across eight perception tasks from ego-centric views. We also introduce vision-language instruction tuning datasets, TB-100k and TB-250k, along with simple yet effective baselines for the tasks. Through extensive experiments, we show that existing MLLMs underperform in these tasks, with even a powerful model like GPT-4o achieving less than 35\% accuracy on average. In contrast, when fine-tuned with TB-100k or TB-250k, our baseline models achieve average accuracy up to 85\%, significantly enhancing performance on the tasks. Additionally, we demonstrate performance transfer by co-training TB-100k with another traffic dataset, leading to improved performance on the latter. Overall, this study represents a step forward by introducing a comprehensive benchmark, high-quality datasets, and baselines, thus supporting the gradual integration of MLLMs into the perception, prediction, and planning stages of AD.
\end{abstract}

%

\section{Introduction}
\label{sec:intro}

\begin{figure}[ht!]
\centering
\includegraphics[width=0.85\linewidth]{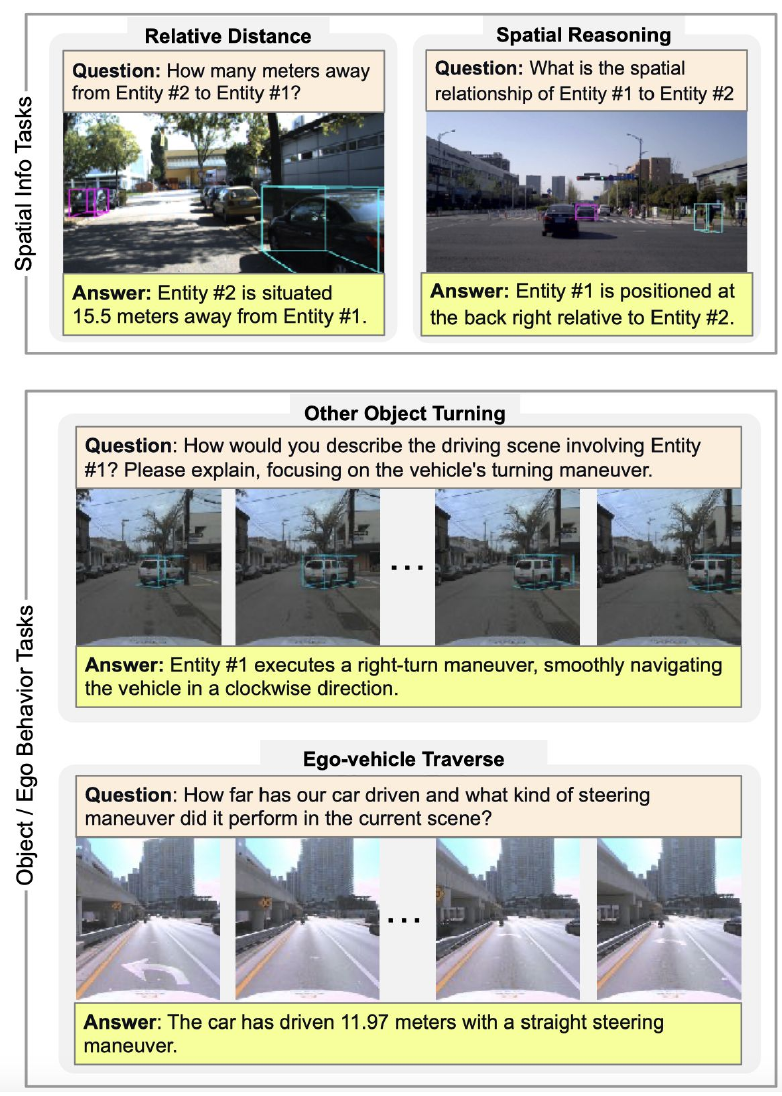}
\caption[Caption for LOF]{Examples of four tasks from TB-Bench; additional task examples are provided in the supplementary material.}
\vspace*{-0.5cm}
\label{fig:benchmark}
\end{figure}

\qt{The application of MLLMs to Autonomous Driving (AD) has gained increasing attention, particularly for predicting risks and planning actions based on images or videos from in-vehicle cameras.
Notably, MLLMs have demonstrated their effectiveness in the international competitions like Autonomous Grand Challenge \cite{renz2024carllava} and in specific tasks such as traffic sign detection \cite{kyberE2E}. However, two major challenges remain.}


\qt{First, current MLLMs, ranging from proprietary models like GPT-4o \cite{achiam2023gpt} and Gemini \cite{anil2023gemini} to open-source models like LLaVA \cite{liu2023visual_llava}, are not optimized for dashcam images or traffic scenes. These models are primarily trained on vast amounts of web-based text and image-text pairs, with minimal traffic-specific data, limiting their effectiveness in AD scenarios. To improve the generalizability of MLLMs, incorporating high-quality domain-specific datasets into the pre-training data is crucial, as shown in \cite{li2024llava_interleave,zhang2024longva}}.

\qt{Second, it lacks a dedicated benchmark for evaluating MLLMs' abilities in spatiotemporal understanding tasks, given their capabilities in vision-centric tasks are still developing. While these models are designed to handle diverse vision-language tasks, they struggle with complex visual understanding, such as spatial reasoning and object relationships \cite{DriveVLM}. 
Even in common domains unrelated to traffic scenes, there are insufficient benchmarks (e.g., Cambrian-1 \cite{tong2024cambrian}, V-star benchmark \cite{wu2024v_vstar}).
Given that AD requires sophisticated geometric and spatiotemporal understanding to capture the dynamic interactions between the vehicle and other entities, high-quality and dedicated benchmarks are much needed. 
}

\qt{While MLLMs are increasingly applied in AD, the aforementioned challenges remain insufficiently addressed. Recent research has primarily focused on using pretrained MLLMs derived from web data for specific AD tasks, without thoroughly investigating these challenges. Another issue is to determine which AD tasks across different stages and levels should be addressed by MLLMs. Within the ``perception, prediction, and planning'' framework, the question becomes: which stages should MLLMs handle?}



\qt{Focusing solely on the perception stage and its associated tasks, it may not necessarily appear optimal to use MLLMs. Established technologies like LIDAR and CV methods such as object detection and visual odometry can accurately capture the vehicle's position and spatiotemporal relationships in Euclidean space.}

\qt{However, when considering the use of MLLMs (or LLMs) in later stages, relying on such ``Euclidean geometrically accurate'' information in the earlier stage might not be optimal. Firstly, it is unclear how to input this information into LLMs. Moreover, achieving advanced understanding in later stages may require information representations specifically suited for MLLMs to perform higher-level tasks. This suggests the need for MLLM involvement from the perception stage.}

\qt{This study adopts this perspective, aiming to involve MLLMs in the perception stage by expressing spatiotemporal traffic scene tasks in natural language text.} The underlying conjecture, as mentioned above, is that this approach could be important for prediction and planning in later stages by MLLMs. 

\qt{To address the challenge of {lacking dedicated benchmarks}, we introduce TB-Bench, one of the first comprehensive benchmarks specifically designed to evaluate MLLM's understanding of traffic behaviors. 
This benchmark assesses MLLM's capabilities to perform perception tasks based on dashcam images or videos from the ego-centric views of vehicles, including determining the spatial position or orientation of other vehicles and interpreting the behaviors of both ego-vehicles and surrounding traffic. Compared to existing benchmarks, TB-Bench encompasses a wider array of eight distinct perception tasks,\footnote{Eight proposed tasks: Relative Distance, Spatial Reasoning, Orientation Reasoning, Other Lane to Ego, Other Lane Changing, Other Turning, Ego Turning, and Ego Traverse Distance.} each corresponding to a typical driver maneuver. Figure \ref{fig:benchmark} shows examples of several tasks. To ensure consistent evaluation across a diverse range of MLLMs, we employ a straightforward protocol. Specifically, we pair questions with images or video clips, requiring an MLLM to respond in plain text. Performance of the MLLM is then assessed by measuring response accuracy.}

\qt{To address the challenge of insufficient training data for AD perception tasks, we introduce a high-quality dataset focused on traffic behavior understanding from ego-centric views. This dataset aligns with the task design of TB-Bench and is used for vision-language instruction tuning (VLIT) of MLLMs. We generate high-quality question-and-answer pairs using samples from established datasets such as KITTI, ONCE, and Argoverse 2. In total, we create \textbf{TB-Bench} comprising 2,000 manually constructed samples, along with two versions of training datasets: \textbf{TB-250k} containing 250,000 samples, and \textbf{TB-100k} (a more balanced version).}
\qt{In addition to evaluating existing MLLMs, we introduce a generic framework that serves as a strong baseline for our tasks, consisting of three standard components: a pretrained vision encoder, a multi-modal connector, and a pretrained LLM. The vision encoder extracts visual representations from inputs with varying number of frames, while the connector projects these embeddings into the LLM's embedding space, finally the LLM generates task-specific responses on our benchmark. This lightweight model is designed for efficient fine-tuning on our proposed dataset(s).}

\qt{Using TB-Bench to evaluate popular proprietary models (GPT-4o and Gemini) and various state-of-the-art open-source MLLMs (LLaVA, Bunny, and InternVL), we find that none of these models excels across all traffic behavior understanding tasks. On average, the open-source models underperform random guessing, while proprietary models achieve only slightly better results, with average accuracy below 35\%.
In contrast, when fine-tuned on TB-100k or TB-250k, our proposed baseline models demonstrate strong performance across all tasks, with average accuracy ranging from 77\% to 85\%. This highlights the effectiveness of our dataset in enhancing MLLM traffic behavior understanding.}

Overall, our contributions are {\color{black}fourfold}: 1) we introduce TB-Bench, a benchmark for assessing MLLMs on eight perception tasks of traffic behavior understanding; 2) we present the VLIT datasets (TB-100k and TB-250k) for the tasks, along with a generic baseline; 3) we conduct extensive experiments demonstrating the performance gap between existing MLLMs and the fine-tuned baselines; {\color{black}and 4) we show that our VLIT dataset, i.e., TB-100k, can be used as part of a co-training dataset to generalize to other driving benchmarks, such as BDD-X \cite{BDD-X}.}

\begin{table*}[ht!]
    \centering
    \caption{
    Summary of existing studies and benchmarks across AD tasks
    (brackets indicate tasks involving planning). 
    }
    \begin{minipage}[t]{0.7\textwidth}
        \vspace{0pt}
            \resizebox{1.0\columnwidth}{!}{
                \begin{tabular}{
                    >{\arraybackslash}p{0.4\textwidth}
                    >{\arraybackslash}p{0.3\textwidth}
                    >{\arraybackslash}p{0.5\textwidth}
                    }
                \toprule
                    \multicolumn{1}{l}{\multirow{1}{*}{\textbf{Benchmarking}}} & 
                    \multicolumn{1}{l}{\multirow{1}{*}{\textbf{Visual Data Modality}}} & 
                    \multicolumn{1}{l}{\multirow{1}{*}{\textbf{ \qt{Perception (Planning) Tasks}}}} \\
                \midrule
                    \rowcolor{gray!10}\textbf{Standalone Task in AD} & & \\
                    DRAMA \cite{DRAMA} & single-image & PER, REA \\
                    Rank2Tell \cite{rank2tell} & single-image & PER, REA, LANE, TLS \\
                    BDD-X \cite{BDD-X} & multi-frame & PER, (AC) \\
                    BDD-OIA \cite{BDD-OIA} & single-image & PER \\
                    TrafficQA \cite{TrafficQA} & multi-frame & PER, PRED, REA \\
                    LingoQA \cite{marcu2023lingoqa} & multi-frame & PER, PRED, REA \\
                    NuScenes-QA \cite{nuscene-qa} & multi-view & OBJ, SP \\
                    NuScenes-MQA \cite{Inoue_2024_WACV} & multi-view & OBJ, \textbf{RD}, OD \\
                    MAPLM-QA \cite{cao2024maplm} & multi-view, BEV-image & LANE \\
                    DriveLM \cite{sima2023drivelm} & single-image & PER, PRED, (PLAN) \\
                    \hdashline
                    \rowcolor{gray!10}\textbf{Benchmark} & & \\
                    SpatialRGPT \cite{cheng2024spatialrgpt} & single-image & \textbf{RD},  \textbf{SR}, \textbf{OR} \\
                    SEED \cite{li2023seed} & multi-image, multi-frame & PER, PRED, REA, AR \\
                    MVBench \cite{li2024mvbench} & multi-frame & PER, PRED, REA, LOC, AR \\
                    MME \cite{Fu2023MMEAC} & single-image & PER, PRED \\
                    MMMU \cite{yue2023mmmu} & multi-image & PER, REA, KNOW \\
                    ELM \cite{zhou2024embodied} & multi-frame & PER, PRED, TLS, OD, OT, AR, (PLAN) \\
                    Cambrian-1 \cite{tong2024cambrian} & single-image & \textbf{RD}, \textbf{SR}, D  \\
                    OpenEQA \cite{majumdar2023openeqa} & multi-frame & OBJ, \textbf{SR}, KNOW, LOC, REA \\
                    \rowcolor{blue!10}\textbf{TB-Bench (Ours)} & single-image, multi-frame & \textbf{RD}, \textbf{SR}, \textbf{OR, EGO-LANE, OBJ-LANE, OBJ-TURN, EGO-TURN, EGO-TRA} \\
                \bottomrule
            \end{tabular}
            }
    \end{minipage}%
    \hfill
    \begin{minipage}[t]{0.3\textwidth}
        \vspace{0pt}
            \centering
            \footnotesize
            \resizebox{0.988\columnwidth}{!}{
                \begin{tabular}{
                    >{\arraybackslash}p{0.32\textwidth}
                    >{\arraybackslash}p{0.68\textwidth}
                    }
                \hline
                \toprule
                \textbf{Abbreviation} & \textbf{Meaning} \\
                \midrule
                OD & 2D \& 3D Object Detection \\ 
                OT & 2D \& 3D Object Tracking \\ 
                D & Depth Estimation \\
                OBJ & Object Existence, Class, etc. \\
                KNOW & World Knowledge \\
                LOC & Location or Coordinate \\
                LANE & Road, Lane, Intersection, etc \\ 
                PER & General Perception \\
                PRED & General Prediction \\ 
                PLAN & General Planning \\
                REA & General Reasoning \\
                TLS & Traffic Light or Sign \\
                AC & Action Category \\
                AR & General Action Recognition \\
                \textbf{RD} & Relative Distance \\
                \textbf{SR} & Spatial Reasoning \\
                \textbf{OR} & Orientation Reasoning \\
                \textbf{EGO-LANE} & Other Lane to Ego-vehicle \\
                \textbf{OBJ-LANE} & Other Lane Changing \\
                \textbf{OBJ-TURN} & Other Turning \\
                \textbf{EGO-TURN} & Ego Turning \\
                \textbf{EGO-TRA} & Ego Traverse Distance \\
                \bottomrule
            \end{tabular}
        }
    \end{minipage}%
    \label{tab:AD_tasks}
\vspace*{-0.4cm}
\end{table*}

{\color{purple}
}

\section{Related Work}

\qt{A summary of existing studies and benchmarks across various AD tasks is presented in Table \ref{tab:AD_tasks}.}
\vspace{-0.2cm}



\subsection{Autonomous Driving Tasks}
The majority of evaluations in the AD field are focused on either end-to-end driving systems, open-loop planning, or standalone task schemes, such as single-round visual question answering (VQA) or captioning. Traditionally, the AD framework consists of perception, prediction, and planning tasks \cite{Reason2Drive}, although slight variations exist, i.e., predicting intention-level outputs instead of trajectories \cite{DriveVLM}.

Generally, perception tasks in end-to-end driving systems are mainly auxiliary tasks, consisting of all available supervision signals 
provided based on the data source. For example, NuScene \cite{Caesar2019nuScenesAM} provides BEV information, segmentation labels, and more. Consequently, multi-task learning is applied to these tasks, such as object detection, tracking, and segmentation.
This approach is consistent across recent similar AD planning datasets, whether in open-loop or simulation scenarios. Occasionally, pretrained VL models are utilized to enhance these modules.

Other popular traffic planning datasets are KITTI \cite{KITTI}, ONCE \cite{ONCE}, Waymo Open \cite{Sun2019ScalabilityIP}, and Argoverse2 \cite{Argoverse2}, which are inherently similar to NuScene in characteristics.




Pretrained VL models are commonly known for their excellence in scene understanding, details, and visual cues. Still, it shows limitations in spatial grounding and reasoning \cite{DriveVLM}. In detail, most standalone task schemes focus on perception tasks, which include general event VQA \cite{TrafficQA, marcu2023lingoqa}, environment and weather conditions, traffic signals, and lane information \cite{wang2023openlanev2, cao2024maplm}. These tasks also encompass critical object detection \cite{DRAMA, rank2tell} or tracking in various forms, such as bounding box coordinates \cite{DriveVLM}, region proposals \cite{talk2car, BDD-OIA}, 2D \cite{wu2023referring}, and 3D \cite{wu2023language} language-guided object tracking, as well as scene analysis that includes attributes or motion of objects like size, position, direction, distance, spatial position relationships \cite{nuscene-qa}, and orientation \cite{cheng2024spatialrgpt}. In particular, there is a comprehensive task for driving with language that integrates all aspects of perception, prediction, and planning in a VQA format \cite{sima2023drivelm}. In the prediction tasks, all previous perception inputs are used to predict the object’s future trajectory, such as parking or moving, and interactions with the ego-vehicle. In the planning stage, it involves combining prior information to generate actions, decision descriptions \cite{BDD-OIA}, and trajectory waypoints \cite{DriveVLM, sima2023drivelm}. 

\subsection{MLLMs and Benchmarks}

VL pre-training and foundation models started with learning from a broader source of supervision, specifically raw text at an internet scale \cite{radford2021learning}, enabling zero-shot transfer of the model to downstream tasks. Notably, approaches attempting to connect VL pre-training to existing LLMs, referred to as MLLMs \cite{Li2023BLIP2BL}, enable capabilities similar to those of LLMs, such as image-to-text generation, improved via instruction tuning and in-context learning capabilities. Current frontier families of MLLMs, such as LLaVA \cite{liu2023visual_llava}, VILA \cite{lin2024vila}, and InternVL \cite{Chen_2024_CVPR}, utilize a similar architectural paradigm: vision encoder, multi-modal projector, and LLM connected in sequence. Despite some early work attempting resampler techniques like Q-Former \cite{dai2023instructblip}, all state-of-the-art models use simpler linear layers with scaling to higher resolutions, focusing on higher quality VLIT instead. Another line of studies works on lightweight versions of MLLMs, optimizing for more informative, condensed training data and design choices \cite{Bunny_He2024EfficientML, imp2024}. The latest MLLMs focus on simultaneously tackling multi-image, multi-frame (video), multi-view (3D), and multi-patch (single-image) scenarios, which show emergent capabilities and enhance overall performance \cite{li2024llava_interleave}. Nevertheless, it is a standard paradigm for MLLMs to evaluate on multiple general benchmarks, aiming to achieve overall performance.

The existing benchmarks, which refer to MLLM benchmarks, aim to comprehensively evaluate various dimensions, but there is no standardized taxonomy for benchmark design. General benchmarks in the VL space started with simple perception-oriented tasks \cite{Fu2023MMEAC}, followed by multi-frame benchmarks \cite{li2023seed, li2024mvbench} with action recognition and VL knowledge-based reasoning \cite{yue2023mmmu}. Spatial or vision-centric benchmarks \cite{tong2024cambrian, cheng2024spatialrgpt} are becoming more relevant to address previously claimed weaknesses. Then, specialized benchmarks gained more attention, introducing tasks from different domains, such as robotics \cite{majumdar2023openeqa} and AD \cite{sima2023drivelm}. 
\qt{In this case, there is still a lack of studies covering simple yet very important skills and behaviors in the AD context.} 




\section{Benchmark Design} \label{benchmark}



\noindent
\qt{TB-Bench is created to fill the benchmark gap in evaluating MLLMs for AD, providing a specialized benchmark that rigorously tests their capability to understand complex traffic behaviors from an ego-centric perspective.}

\subsection{Task Design}
We generate question-and-answer pairs in a VQA format, where the model takes an image or video paired with a question as input and produces a corresponding answer. Both the question and answer are expressed in a single sentence of free-form text.

To achieve the above goal, we consider multiple types of Q\&A pairs, each linked to a specific driver's maneuver behavior. We refer to the Pre-crash Scenarios typology from the National Automotive Sampling System (NASS) variables \cite{nhtsa2007precrash}, which are also utilized in the CARLA simulator \cite{dosovitskiy2017carla}. This typology includes a total of 65 pre-crash scenarios, categorized into nine accident types\footnote{The accident types are Animal, Off-road, Pedalcyclist, Pedestrian, Backing, Lane Change, Opposite Direction, Rear-end, and Crossing-paths.}. Each scenario is described in the format of `{an accident type: a detailed scenario}.' \qt{For example, the `lane change' accident type includes scenarios like `one vehicle passing while another is turning.' See the supp. material for the full list of scenarios.}

\qt{Focusing on typical maneuver behaviors derived from NASS scenarios, we have identified eight distinct Q\&A types, referred to as `tasks,' as shown in Table~\ref{tab:abstract_concepts}.}
Some tasks require numerical outputs (e.g., `distance in meters'), while others require discrete classes (e.g., `back,' `back left,' etc.). It is important to note that the models are expected to provide these outputs in their natural language responses. 
\qt{Fig. \ref{fig:benchmark} presents examples for four of the eight tasks, each of which consists of input image(s) accompanied by a question and a ground-truth answer.}
The visual input is either a single image or multiple images (up to eight), depending on the task, as will be explained later.

\begin{table}[t]
    \caption{
    \textbf{Tasks and Concepts Addressed in Each.} 
    `Classes' column indicates the types of outputs, i.e., the number of discrete classes or numerical outputs (indicated by $\mathcal{R}$
    ); `Orientation Reasoning' task contains both output types. 
    } 
    \centering
    \resizebox{1.0\columnwidth}{!}{
        \begin{tabular}{
            >{\arraybackslash}p{0.40\columnwidth}
            >{\arraybackslash}p{0.54\columnwidth}
            >{\centering\arraybackslash}p{0.15\columnwidth}
        }
        \toprule
        \textbf{Task Type} & \textbf{Abstract Concepts} & \textbf{
        Classes} \\
        \midrule
        \rowcolor{gray!10}\textbf{Spatial Information:} & & \\
        \hspace{1mm} Relative Distance & distance in meters & $\mathcal{R}$ \\
        \hspace{1mm} Spatial Reasoning & back, back left, back right, front, front left, front right & 6 \\
        \hspace{1mm} Orientation Reasoning & opposite, perpendicular, similar, and degrees & 3/$\mathcal{R}$ \\
        \midrule
        \rowcolor{gray!10}\textbf{Object Behavior:} & & \\
        \hspace{1mm} \parbox[t]{3cm}{Other Lane to Ego-Vehicle} & front lane, front left lane, front right lane, oncoming traffic lane & 4 \\
        \hspace{1mm} Other Lane Changing & left lane change, no change, right lane change & 3 \\
        \hspace{1mm} Other Turning & go straight, left turn, right turn & 3 \\
        \midrule
        \rowcolor{gray!10} 
        \textbf{Ego Behavior:} & & \\
        \hspace{1mm} Ego Turning & go straight, left turn, right turn & 3 \\
        \hspace{1mm} Ego Traverse Distance & distance traveled in meters & $\mathcal{R}$ \\
        \bottomrule
    \end{tabular}
    }
    \label{tab:abstract_concepts}
\vspace{-0.5cm}
\end{table}



\subsection{Referencing Entities}
Some tasks require the model to determine the spatial position or orientation of other vehicles, as shown in Fig.~\ref{fig:benchmark}. When multiple vehicles are present in a scene, it is essential to distinguish between them in both the questions and answers. One approach is to describe the vehicle by its attributes, such as ``black compact sedan,'' but this can pose challenges in ensuring the model accurately identifies and differentiates similar objects using such descriptions. To avoid these complications and focus on evaluating the model's spatial understanding, we 
\qt{label} each target traffic entity as `Entity \#$n$' in the questions and answers, where $n$ corresponds to its index in the input image(s); see examples in the upper part of Fig.~\ref{fig:benchmark}. To identify these entities, we draw colored three-dimensional bounding boxes (BBs) directly in the input image(s), using a consistent color for each entity index $n$ throughout the dataset. Specifically, we use cyan and magenta BBs for `Entity \#$1$' and `Entity \#$2$,' respectively. Our dataset includes up to two entities per scene, i.e., $n = 1$ or $2$. An additional advantage of this method is that it requires minimal instruction tuning or even no extra learning for MLLMs to adapt. Furthermore, it is compatible with multi-view, multi-frame, and multi-scale modalities, as demonstrated in AnyRes \cite{liu2024improved_anyres}, UniRes \cite{zhang2024longva}, and Interleave \cite{li2024llava_interleave}.

\subsection{Evaluation}

Our benchmark requires MLLMs to generate plain text outputs. Since the goal is to evaluate the \qt{spatiotemporal} understanding capabilities 
of MLLMs, the accuracy of their outputs should be assessed using methods tailored to this requirement. 

The questions in the dataset are broadly classified into two categories based on the type of answers expected. One category includes questions about positional relationships or orientation, with typical answers like ``positioned at the back right'' or ``a right-turn maneuver.'' The other category involves questions requiring numerical answers, such as ``is situated 15.53 meters away.''

For the first category of Q\&A, keywords are manually selected for each task or ground truth answer, and their presence in the output text is identified using rule-based methods (i.e., regular expressions). For the second category, the predicted value is compared to the correct answer, and if the difference falls within a specified range, the prediction is considered correct; otherwise, it is deemed incorrect. In the experiments, thresholds are set such that a difference within 25\% of the correct value is considered acceptable for distance, and a difference within 15 degrees is acceptable for angle. \qt{Refer to the supp. material for more details.}

\section{Generation of VQA Data}
\begin{figure*}[t]
\centering
\includegraphics[width=0.85\linewidth]{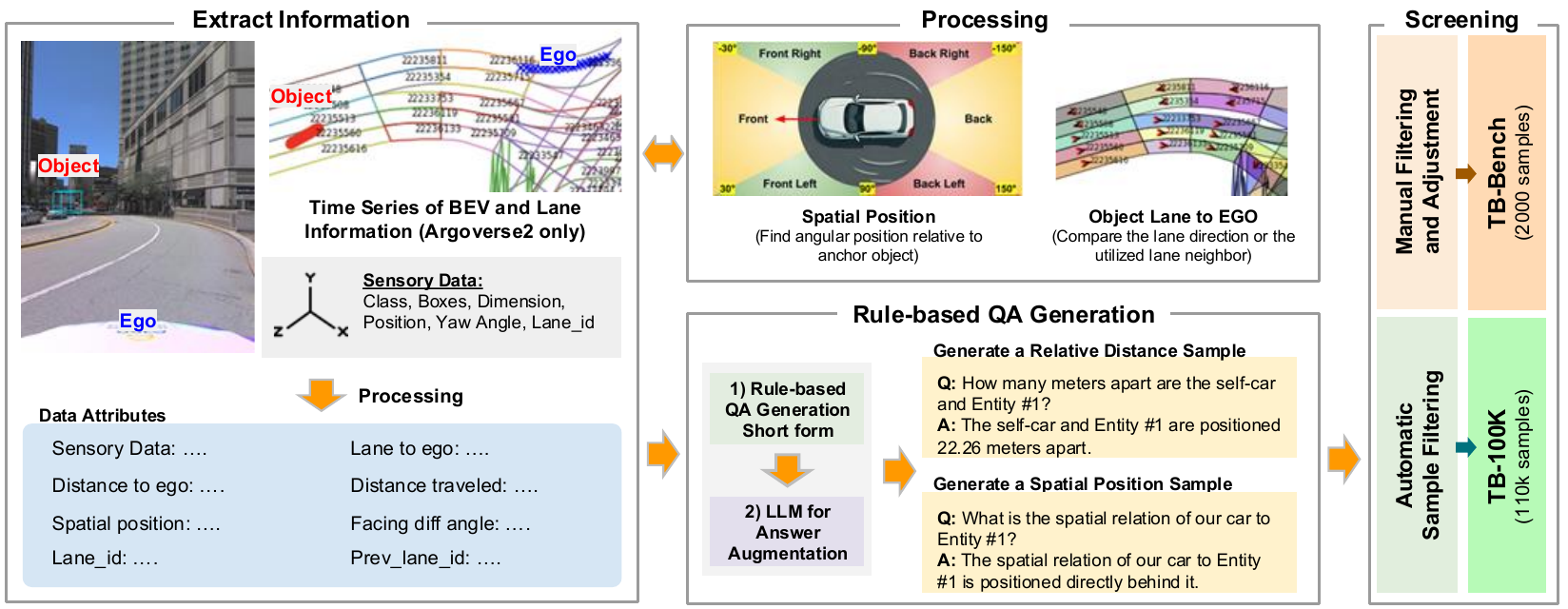}
\caption{
\qt{
\textbf{Overview of Data Generation Pipeline}.
Left: Sensory data is processed into higher-level attributes. Middle-Top: Spatial positioning and lane orientation relative to the ego-vehicle are determined. Middle-Bottom: Q\&A samples are generated using rules and LLM augmentation. Right: Data is filtered and refined for the final dataset.
}
}
\vspace*{-3mm}
\label{fig:data_generaion}
\end{figure*}
\subsection{Outline}

To generate Q\&A pairs for the eight tasks mentioned, we repurpose existing datasets, specifically KITTI \cite{Geiger2012CVPR}, ONCE \cite{ONCE}, and Argoverse2 \cite{Argoverse2}. These datasets are originally designed for 
studying object detection, localization, and tracking in three-dimensional space, providing detailed three-dimensional geometry of traffic entities. KITTI and ONCE, in particular, offer object class information and 3D bounding boxes for each traffic entity, including their position, dimensions, and yaw angle. Argoverse2 further enriches this with lane information relative to the ego vehicle.

To align with the task design mentioned (Table \ref{tab:abstract_concepts}), the quantities provided by these datasets, mostly represented in the Euclidean space, are converted into abstract concepts, such as six discrete angles between two vehicles (e.g., front right, back left, etc.), lanes relative to the ego-car (i.e., front left lane, oncoming lane) and lane changing.

For the first three tasks—`Relative Distance,' `Spatial Reasoning,' and `Orientation Reasoning'—we generate Q\&A pairs using samples from KITTI and ONCE, as these tasks do not require lane information from the ego vehicle or others. Since these tasks can be performed using a single image, we utilize a static dashcam image as the visual input. For the remaining tasks—`Other Lane to Ego,' `Other Lane Changing,' `Other Turning,' `Ego Turning,' and `Ego Traverse Distance'—which require lane information \kt{and a multi-frame source}, we generate Q\&A pairs using Argoverse2. Given that these tasks involve temporal changes, we extract eight image frames from the `long scenario' sequences in the dataset for each Q\&A pair\footnote{
\qt{Each `long scenario' sequence in the dataset is 15 seconds long. From these, we extract 1.6-second clips, consisting of eight images captured at 0.2-second intervals.}
}, using these sequences as the visual input for models.

After generating the data automatically, we conduct a manual screening process. Based on the extent of screening, the data is organized into three distinct datasets. One dataset, comprising 2,000 samples, is designated for evaluation purposes, which we will refer to as `benchmark' in this paper. These samples undergone thorough manual inspection, removing low-quality samples and ensuring an equal number (i.e., 250) of samples per task. The remaining two datasets are intended for model training: the first, TB-250k, contains 250,000 samples; the second, TB-100k, includes \kt{over} 100,000 samples that have been filtered to balance the number of samples per task. Table \ref{tab:data_samples} summarizes the overall statistics of these datasets. 

\begin{table}[t]
    \caption{\qt{
    \textbf{Statistics of TB-Bench, TB-100k, and TB-250K}.  
    Source datasets: K (KITTI), O (ONCE), Arv2 (Argoverse2).
    }}
    \vspace{-0.5cm}
    \centering
    \resizebox{0.45\textwidth}{!}{%
    \begin{tabular}{
            >{\arraybackslash}p{0.43\columnwidth} 
            >{\centering\arraybackslash}p{0.15\columnwidth}
            >{\centering\arraybackslash}p{0.10\columnwidth}
            >{\centering\arraybackslash}p{0.08\columnwidth}
            >{\centering\arraybackslash}p{0.08\columnwidth}
            }
        \toprule
            \textbf{Task Type} & \textbf{Sources/\break Frames} &\textbf{TB-Bench} & \textbf{TB-250k} & \textbf{TB-100k} \\ 
        \midrule
            \rowcolor{gray!10}\textbf{Spatial Information:} & & & & \\
            \hspace{1mm} Relative Distance & [K, O]/1 & 250 & 35k & 10k \\
            \hspace{1mm} Spatial Reasoning & [K, O]/1 & 250 & 70k & 30k \\
            \hspace{1mm} Orientation Reasoning & [K, O]/1 & 250 & 70k & 30k \\
        \midrule
            \rowcolor{gray!10}\textbf{Object Behavior:} & & & & \\
            \hspace{1mm} Other Lane to Ego  & [Arv2]/8  & 250 & 50k & 20k \\
            \hspace{1mm} Other Lane Changing & [Arv2]/8  & 250 & 1.5k & 1.5k \\
            \hspace{1mm} Other Turning & [Arv2]/8  & 250 & 1.5k& 1.5k \\
        \midrule
            \rowcolor{gray!10}\textbf{Ego Behavior:} & & & & \\
            \hspace{1mm} Ego Turning & [Arv2]/8 & 250 & 1.5k & 1.5k \\
            \hspace{1mm} Ego Traverse Distance & [Arv2]/8 & 250 & 25k & 15.5k \\
        \midrule
            \textbf{Total} & & 2000 & 254k & 110k \\
        \bottomrule
    \end{tabular}
    }
    \label{tab:data_samples}
\vspace{-0.5cm}
\end{table}

\begin{figure*}[t]
\centering
\includegraphics[width=0.84\linewidth]{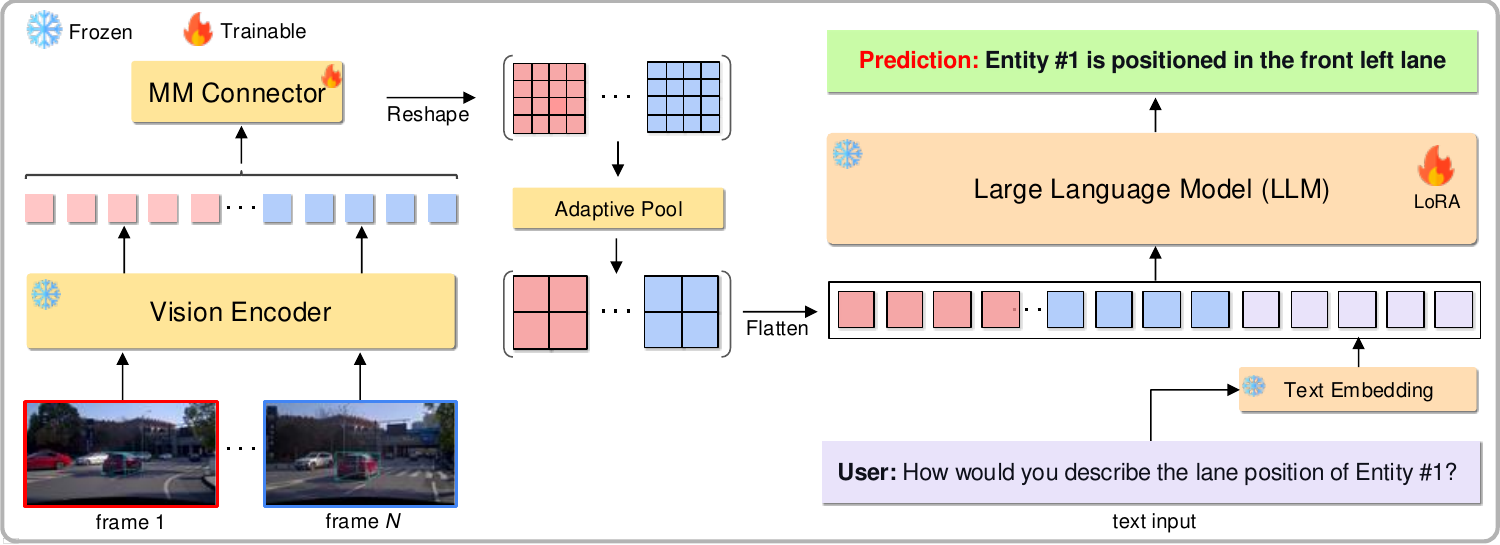}
\caption{\qt{The overall architecture of our baseline framework.}}
\label{fig:proposed_method}
\vspace*{-0.4cm}
\end{figure*}

\subsection{Details of the Pipeline}
The Q\&A pairs are generated automatically, with manual inspection following the automated process. The only exception is the `Other Lane Changing' task, \kt{where we manually generate Q\&A pairs due to noisy lane information at intersections}.
Figure \ref{fig:data_generaion} illustrates the pipeline used for generating Q\&A pairs from these datasets.

The process unfolds as follows: The input to the pipeline is a single sample from the datasets, which could be either a static image with a set of entity attributes from KITTI/ONCE or a list of sequences with similar data from Argoverse2. The pipeline begins by extracting key information from the input, as depicted in the left panel of Fig.~\ref{fig:data_generaion}. This is followed by a processing step shown in the middle-top panel of Fig.~\ref{fig:data_generaion}, where spatial positions and facing angles relative to an anchor object are calculated. Additionally, the `lane to ego' task identifies on which side the entity is located relative to the ego vehicle. For turning behaviors, we record the accumulated turning angle of each object to determine its recent motion. For lane changes, a flag is recorded if there are changes in lane\_id compared to the previous step. Similarly, all sensor numerical ground truth data—such as position, dimension, and angle of all entities—are processed into attributed data, such as distance to ego and spatial position. 

Finally, a rule-based process, depicted in the middle-bottom panel of Fig.~\ref{fig:data_generaion}, is triggered to identify the corresponding task and generate Q\&A pairs. Further details can be found in the supplementary material.

In the next phase, a rule-based system generates QA samples from processed data attributes. This depends on the type of task, i.e., tasks aside from lane change and turning behavior can be created based on any frame, without necessarily needing an event to trigger it. Thus, they naturally have more data samples generated. After this, the rule-based QA is generated with simple short answers, such as `oncoming lane' or `turn left.' Then, it is augmented to be a more complex sentence using text-only information with an LLM; we used Microsoft-Phi3-medium \cite{abdin2024phi}.

\section{Baseline Framework}
\qt{We present a generic framework that serves as a strong baseline for our tasks, comprising three standard components: a vision encoder, a multi-modal connector (a two-layer MLP), and an LLM. The vision encoder extracts visual representations from input frames, the multi-modal connector projects these representations into the LLM's embedding space, and finally the LLM generates a response based on the given question and visual embeddings. Figure~\ref{fig:proposed_method} illustrates the architecture of our framework.} 

\qt{We now explain how to adaptively extract visual representations from varying numbers of frames and input them into the LLM. Given $N$ frames of $H \times W$ having color-coded bounding boxes, the vision encoder processes each frame individually to produce $N$ visual representations of size $[H/p \times W/p, C]$, where $p$ is the patch size and $C$ is the embedding dimension of the encoder. These visual representations are then projected into the LLM's embedding space of $D$ using the multi-modal connector, resulting in $N$ visual embeddings of size $[H/p \times W/p, D]$.}

\qt{Inputting all visual embeddings of $N$ frames into the LLM can be computationally expensive. To address this, we sample spatially a subset of these visual embeddings per frame. Specifically, we apply adaptive average pooling to reduce each frame's embeddings, from $[H/p \times W/p, D]$ to $[k=h\times w, D]$, where $k \ll H/p \times W/p$. The value of $k$ is determined as a hyperparameter. The sampled embeddings from all $N$ frames are then reshaped and concatenated, preserving spatial and temporal order, which yields final visual embeddings of size $[N \times k, D]$ that are passed into the LLM along with the textual embeddings.}

\qt{To process text input, we tokenize the question and its ground-truth response, converting them into textual embeddings. These are then combined with the visual embeddings and input into the LLM. We train the model by minimizing cross-entropy loss on the response token predictions. During inference, only the question is used as text input.}

\section{Experiments}
\begin{table*}[t]
\centering
\caption{
Results of compared methods on TB-Bench are reported in accuracy (\%), where higher indicates better performance. Random guess$^\dagger$ results are considered zero. $^\star$In-context learning for single-frame tasks uses three in-context examples, while multi-frame tasks use one. Huggingface and API names are used for easy reference.
}

\resizebox{0.98\textwidth}{!}{%
\begin{tabular}
    {
    >{\arraybackslash}p{0.26\textwidth}
    >{\centering\arraybackslash}p{0.06\textwidth}
    >{\centering\arraybackslash}p{0.06\textwidth}
    >{\centering\arraybackslash}p{0.06\textwidth}
    >{\centering\arraybackslash}p{0.11\textwidth}
    >{\centering\arraybackslash}p{0.11\textwidth}
    >{\centering\arraybackslash}p{0.11\textwidth}
    >{\centering\arraybackslash}p{0.11\textwidth}
    >{\centering\arraybackslash}p{0.11\textwidth}
    >{\centering\arraybackslash}p{0.06\textwidth}
    }
\toprule
    \multicolumn{1}{c}{\multirow{3}{*}{\Large\textbf{Model}}} & 
    \multicolumn{9}{c}{\multirow{2}{*}{\Large\textbf{TB-Bench Tasks}}}
    \\
    \\
    & \multirow{1}{*}{\textbf{RD} $\uparrow$} 
    & \multirow{1}{*}{\textbf{SR} $\uparrow$} 
    & \multirow{1}{*}{\textbf{OR} $\uparrow$} 
    & \multirow{1}{*}{\textbf{EGO-LANE} $\uparrow$} 
    & \multirow{1}{*}{\textbf{OBJ-LANE} $\uparrow$}
    & \multirow{1}{*}{\textbf{OBJ-TURN} $\uparrow$}
    & \multirow{1}{*}{\textbf{EGO-TURN} $\uparrow$}
    & \multirow{1}{*}{\textbf{EGO-TRA} $\uparrow$}
    & \multirow{1}{*}{\textbf{Avg.} $\uparrow$}
    \\
\midrule
    Random$^\dagger$ & 0.0 & 16.7 & 17.1 & 25.0 & 33.3 & 33.3 & 33.3 & 0.0 & 19.8 \\
    \midrule
    \rowcolor{gray!10}\textbf{Zero-shot 
    } & & & & & & & & & \\
    LLaVA-1.5-7B & 10.8 & 16.8 & 28.0 & 28.4 & 20.4 & 23.2 & 16.8 & 0.0 & 18.1 \\
    LLaVA-v1.6-Mistral-7B & 4.0 & 25.6 & 30.8 & 20.4 & 26.0 & 22.4 & 27.2 & 0.0 & 19.6  \\
    LLaVA-NeXT-Video-7B & 3.6 & 0.8 & 13.2 & 10.4 & 18.8 & 22.4 & 30.0 & 0.0 & 12.4  \\
    LLaVA-Interleave-Qwen-7B & 5.6 & 24.8 & 10.8 & 31.6 & 19.2 & 26.8 & 20.4 & 0.0 & 17.4  \\
    Bunny-v1.1-4B & 24.4 & 20.4 & 19.6 & 28.4 & 16.0 & 20.0 & 34.4 & 0.0 & 20.4  \\ 
    Bunny-v1.1-Llama-3-8B-V & 7.6 & 16.4 & 30.0 & 26.8 & 18.4 & 21.6 & 20.0 & 1.2 & 17.8 \\
    InternVL2-8B & 3.6 & 12.0 & 28.0 & 28.4 & 28.0 & 29.2 & 30.4 & 0.4 & 20.0  \\
    Mini-InternVL2-1B-DriveLM & 0.0 & 31.2 & 20.0 & 28.4 & 24.8 & 47.2 & 41.6 & 0.0 & 24.2  \\
    DriveLM-mantis-8B & 0.0 & 34.8 & 23.2 & 30.0 & 57.6 & 50.8 & 48.8 & 0.0 & 30.7 \\
    Gemini-1.5-flash & 21.2 & 16.8 & 22.0 & 34.8 & 48.0 & 23.2 & 27.6 & 4.8 & 24.8  \\
    \kt{GPT-4o-2024-08-06} & 8.4 & 32.0 & 40.8 & 54.4 & 39.6 & 43.2 & 40.4 & 16.0 & 34.4  \\
    \midrule
    \rowcolor{gray!10}\textbf{In-context learning}$^\star$ & & & & & & & & & \\
    LLaVA-Interleave-Qwen-7B & 14.0 & 3.6 & 10.4 & 24.8 & 29.6 & 19.6 & 28.0 & 24.4 & 19.3  \\
    \kt{GPT-4o-2024-08-06} & 32.8 & 38.8 & 36.8 & 60.4 & 51.2 & 38.4 & 46.4 & 22.8 & 40.9  \\
    \midrule
    \rowcolor{gray!10}\textbf{VLIT on TB-100k} & & & & & & & & & \\
    \textbf{Ours} (SigLIP-L-Qwen1.5-0.5B) & 76.4 & 74.4 & 86.8 & 94.0 & 68.8 & 74.8 & 81.2 & 63.2 & 77.5  \\
    \textbf{Ours} (SigLIP-L-Qwen2-0.5B) & 80.4 & 74.8 & 88.8 & 93.6 & 65.2 & 76.4 & 80.0 & 60.4 & 77.5  \\
    \midrule
    \rowcolor{gray!10}\textbf{VLIT on TB-250k} & & & & & & & & & \\
    \textbf{Ours} (SigLIP-L-Qwen1.5-0.5B) & 93.6 & 82.4 & 96.0 & 99.6 & 69.6 & 80.4 & 82.0 & 73.4 & 84.5  \\
    \textbf{Ours} (SigLIP-L-Qwen2-0.5B) & 91.2 & 83.2 & 94.8 & 99.6 & 69.6 & 80.4 & 82.8 & 78.8 & 85.1  \\

\bottomrule
\end{tabular}%
}
\vspace{-0.2cm}
\label{tab:main_result}
\end{table*}

\subsection{Experimental Settings}
\qt{Our proposed framework is compatible with any vision encoder and LLM. In this study, we utilize pretrained SigLIP-L/14 \cite{zhai2023sigmoid_siglip} as the vision encoder and the powerful pretrained Qwen 0.5B, either version 1.5 or 2.0, \cite{qwen,yang2024qwen2} as the LLM, while initializing the parameters of the multi-modal connector randomly. To preserve pretrained LLM capabilities and enable efficient task-specific fine-tuning, we apply LoRA \cite{hu2022lora} with a rank of 64. During training, we freeze the vision encoder and LLM parameters, updating only the parameters of the multi-modal connector and LoRA adapters.}

\qt{For tasks requiring temporal information, the number of frames $N$ is 8; otherwise $N=1$. Each frame is resized to $384\times 384$ as the input to SigLIP-L/14, with the number of sampled visual embeddings k set to $16$ (i.e., $h = w = 4$).}

\qt{We fine-tune our models on either TB-100K or TB-250K, and then report the accuracy on TB-Bench. We use AdamW \cite{loshchilov2017decoupled} with a learning rate of 2e-4 and batch size of 64 for 10 epochs, with learning rate adjusted via a cosine scheduler.}

\subsection{Zero-shot Evaluation for MLLMs}

We report the zero-shot performance of various MLLMs on TB-Bench, including two popular proprietary models (GPT-4o, Gemini 1.5), several SOTA open-source general models, including LLaVA \cite{liu2023visual_llava}, Bunny \cite{Bunny_He2024EfficientML}, and InternVL \cite{Chen_2024_CVPR}, {\color{black}as well as open-source models with traffic domain adaptations trained on DriveLM \cite{sima2023drivelm}, i.e., Mantis \cite{jiang2024mantis} and Mini-InternVL2 \cite{gao2024mini_miniinternvl}}. For class output questions, we use a multi-choice template listing all possible class options, while for numerical output questions, we specify the format, i.e., ``Answer in xx.x meters.'' See the supp. material for more details on the models and the prompt design.



\subsection{Results on TB-Bench}
\qt{ Table \ref{tab:main_result} shows the results of different methods on TB-Bench tasks, categorized into four groups: zero-shot evaluation{\color{black}, in-context learning evaluation}, VLIT on TB-100k, and VLIT on TB-250k.}

In the zero-shot evaluation, although the proprietary models (GPT-4o and Gemini) outperform the open-source models overall, none of them excels across all traffic behavior tasks. {\color{black}Many open-source models underperform random guessing, while traffic domain adaptation models show significantly better performance in certain areas but still lag behind the proprietary models. The proprietary models achieve an average accuracy of less than 35\%.}

{\color{black}
In in-context learning, examples significantly improve performance in specific areas, i.e., numerical outputs.
}

\qt{For baseline models fine-tuned on TB-100k, both with Qwen variants demonstrate strong performance across all tasks, with an average accuracy of 77.5\%. Even the lowest-performing task exceeds 60\% accuracy, showing a significant improvement of over almost 45\% compared to GPT-4o and 57\% over random chance. This underscores the effectiveness of VLIT when a high-quality dataset is available, enhancing traffic behavior understanding of MLLMs.}

\qt{For baseline models fine-tuned on TB-250k, performance improves across all tasks, particularly those with increased data samples. Notably, accuracy in tasks like OBJ-LANE, OBJ-TURN, and EGO-TURN, with the same number of training samples to TB-100k, also benefits from additional samples in other tasks. This suggests that learning from tasks can be transfered to those with limited training data.}

\subsection{Abalation Study}
\begin{table}[htbp]
\setlength{\tabcolsep}{6pt}
\centering
\footnotesize
\caption{Ablation results on (a) vision encoders,  (b) number of visual embeddings per frame, and (c) number of frames.} 
\label{tab:ablations}
\vspace{-0.2cm}
\subfloat[vision encoder]{
\begin{tabular}{cc}
\toprule
Encoder & Acc \\
\midrule
CLIP-L/14 & 72.0 \\
SigL-B/16 & 74.3 \\
SigL-L/14 & 77.5 \\
\bottomrule
\label{tab:ablation:a}
\end{tabular}
\vspace{0.4cm}
}
\subfloat[\# tokens/frame]{
\begin{tabular}{cc}
\toprule
\# tokens/fr & Acc \\
\midrule
4 & 72.7 \\
16 & 77.5 \\
36 & 76.2 \\
\bottomrule
\label{tab:ablation:b}
\end{tabular}
}
\subfloat[\# frames]{
\begin{tabular}{cc}
\toprule
\# frames & Acc \\
\midrule
2 & 72.1 \\
4 & 73.8 \\
8 & 77.5 \\
\bottomrule
\label{tab:ablation:d}
\end{tabular}
}
\vspace{-1cm}
\end{table}

\begin{table*}[ht]
\centering
\small
\caption{ {\color{black}Quantitative results of action tasks on BDD-X test dataset. We provide evaluation results on action description, action justification, and full-text generation (i.e., combining description and justification). `B4' stands for BLEU4.}}
\begin{tabular}{lccccccccc}
\toprule
\multirow{2}{*}{Method} & \multicolumn{3}{c}{Description} & \multicolumn{3}{c}{Justification} & \multicolumn{3}{c}{Full} \\ \cmidrule(lr){2-4} \cmidrule(lr){5-7} \cmidrule(lr){8-10}
 & CIDEr & B4 & ROUGE & CIDEr & B4 & ROUGE & CIDEr & B4 & ROUGE \\
\midrule
\midrule
ours (BDD-X) & 118.6 & 20.0 & 53.8 & 61.3 & 6.9 & 26.1 & 54.2 & 12.0 & 38.4 \\
\midrule
ours (BDD-X + \textbf{TB-100k}) & 121.7 & 20.0 & 54.3 & 60.3 & 6.7 & 26.7 & 53.7 & 11.9 & 38.6 \\
\bottomrule
\end{tabular}
\label{tab:generalization_to_bdd_x_actions}
\end{table*}

\begin{table*}[ht]
\centering
\small
\caption{{\color{black}Quantitative results of control signals prediction on BDD-X test dataset. RMSE denotes the root mean squared error, and $A_\tau$ measures the proportion of test samples with prediction errors less than $\tau$.}}
\begin{tabular}{lcccccccccc}
\toprule
\multirow{2}{*}{Method} & \multicolumn{5}{c}{Speed (m/s)} & \multicolumn{5}{c}{Turning angle (degree)} \\ \cmidrule(lr){2-6} \cmidrule(lr){7-11}
 & RMSE$\downarrow$ & $A_{0.1}\uparrow$ & $A_{0.5}\uparrow$ & $A_{1.0}\uparrow$ & $A_{5.0}\uparrow$ & RMSE$\downarrow$ & $A_{0.1}\uparrow$ & $A_{0.5}\uparrow$ & $A_{1.0}\uparrow$ & $A_{5.0}\uparrow$ \\
\midrule
\midrule
ours (BDD-X) & 1.40 & 26.1 & 55.7 & 75.6 & 98.6 & 11.2 & 44.2 & 62.2 & 71.8 & 89.2 \\
\midrule
ours (BDD-X + \textbf{TB-100k}) & 1.38 & 26.3 & 57.6 & 76.1 & 98.8 & 11.3 & 44.5 & 63.7 & 73.0 & 89.3 \\
\bottomrule
\end{tabular}
\label{tab:generalization_to_bdd_x_controls}
\end{table*}

\qt{We conduct an ablation study to identify which factors enhance performance during fine-tuning, regarding visual inputs to the models. All experiments use the same settings unless noted. The results are summarized in Table~\ref{tab:ablations}.}

\qt{Table \ref{tab:ablation:a} compares different pretrained vision encoders, including CLIP-L/14 \cite{radford2021learning} and SigLIP-B/16 (processing $224 \times 224$ frames). It is seen that the SigLIP encoders outperform the CLIP encoder, with SigLIP-L/14 achieving the highest accuracy.}

\qt{Table \ref{tab:ablation:b} presents the results of using varying numbers of sampled visual embeddings/tokens per frame, $p$ (where $h = w = \sqrt{p}$). We observe that using 16 sampled visual tokens per frame is optimal, and increasing $p$ can degrade performance.}

Finally, we evaluate the impact of varying the number of sampled frames $N$(= 2, 4, 8), on the tasks requiring temporal information. We consistently select the first and last frames, with the remaining $N-2$ frames sampled uniformly in between. As shown in Table \ref{tab:ablation:d}, increasing temporal information significantly boosts performance. For detailed task accuracy and other ablation results, see the supp. material.

{\color{black}
\subsection{Cross-Dataset Generalization}

We conduct additional experiments to demonstate the performance transfer from co-training with perception-stage tasks and planning tasks, to show its improvements on downstream tasks. Specifically, we co-train the TB-100k dataset with the BDD-X dataset and evaluate on the action and control prediction tasks \cite{BDD-X}.

As the BDD-X dataset involves frame index referencing in both the question and answer text annotations, we employed the Mini-InternVL model \cite{gao2024mini_miniinternvl} as the baseline, which formulates frame referencing in a similar manner.

We follow a standard MLLM training regime: Stage 1 focuses on feature alignment, utilizing the pre-trained checkpoint of Mini-InternVL, while Stage 2 involves instruction tuning on the main datasets. In the standalone setting, the main dataset involves tuning with BDD-X for 20 epochs, while in the co-training setting, we tune the mixed BDD-X dataset for 20 epochs and TB-100k for 1 epoch. We apply LoRA \cite{hu2022lora} with a rank of 64. During training, we freeze the vision encoder and LLM parameters, updating only the parameters of the multi-modal connector and LoRA adapters. Overall, training is conducted with a learning rate of 2e-4 and a batch size of 96.

Tables \ref{tab:generalization_to_bdd_x_actions} and \ref{tab:generalization_to_bdd_x_controls} compare the transfer performance between standard training and additional co-training with TB-100k. Notably, beyond differences in task types within the VQA format, the two tables also differ in the type of outputs, i.e., free-form text and numerical outputs.

Table \ref{tab:generalization_to_bdd_x_actions} shows improved performance with the co-training setting in the description data split, which includes annotations about scene perception. However, there are marginal differences in the other splits, which are not directly related to perception tasks.

Table \ref{tab:generalization_to_bdd_x_controls} demonstrates consistent performance improvement with co-training across most metrics, except for the RMSE of the turning angle, which shows a slight decrease.
}

\section{Conclusion}

We have introduced TB-Bench, a comprehensive benchmark that rigorously assesses MLLM performance across eight perception tasks, providing a much-needed standard for spatiotemporal evaluation in AD. Alongside TB-Bench, we have developed the vision-language instruction tuning datasets, TB-100k and TB-250k, which significantly improve MLLM performance when used to fine-tune our baseline models, resulting in substantial gains over existing models. {\color{black} Additionally, our VLIT datasets offer benefits as valuable assets for mixed training datasets in other driving use cases.} Our contributions not only represent incremental progress, but also lay a solid foundation for the further integration of MLLMs into the perception, prediction, and planning stages of AD. These resources are poised to accelerate advancements in the field, supporting the development of more capable and reliable autonomous systems. Please refer to the supplementary material for further discussion on broader impact, limitations, and future work.


\end{document}


\maketitle

\noindent This material includes the following sections:

\begin{itemize}
    \item \textbf{Discussions}: The broader impact, limitations, and future directions of our work.
    \item \textbf{Access Information}: A URL for accessing the benchmark, datasets, and future update.
    \item \textbf{Task Definitions and Dataset Statistics}: A detailed overview of the task definitions and relevant dataset statistics.
    \item \textbf{Data Generation Pipeline}: Insights into the Data Generation Pipeline used in our study.
    \item \textbf{Evaluation Details}: Information on metrics, models, and evaluation methods.
    \item \textbf{Experiments and Results}: Implementation details, quantitative analyses, qualitative results, and ablation studies.
\end{itemize}

\section{Discussions}

\subsection{Broader Impact} 

\qt{This study represents progress in enhancing the capabilities of Multi-Modal Large Language Models (MLLMs) by focusing on a limited set of AD perception tasks. Specifically, we introduce a new benchmark to evaluate MLLMs on understanding diverse traffic behaviors and provide high-quality VLIT datasets that enhance MLLMs' generalizability. We hope this will advance MLLMs' applications in AD, contributing to the development of more robust autonomous driving systems.}



\subsection{Limitations} 
\qt{Firstly, our study utilizes the moderate large language models (Qwen 0.5B series) due to limited computational resources, which can be scaled up as needed.}

\qt{Secondly, we acknowledge the dataset imbalance arising from the natural occurrence of specific autonomous driving behaviors; please refer to Section Dataset Statistics for more details.}

\qt{Lastly, the free-form text output templates in TB-100k and TB-250k are limited for certain tasks. However, we believe that the diversity of images is also important for the model to understand visual concepts. That being said, when combined with other (vision-)language instruction tuning datasets, our datasets still enhance the performance of MLLMs, enabling them to generalize better in traffic domains, particularly in understanding traffic behaviors.}




\subsection{Future Work}
\qt{Future research could expand this work by incorporating a wider range of perception tasks or by exploring subsequent stages, such as prediction and planning.}

\qt{Additionally, an important direction for future investigation is the optimal application of upstream perception tuning sets, including the TB-100k and TB-250k datasets, to relevant downstream traffic tasks. This approach may enhance model performance in real-world applications.}

\qt{Furthermore, integrating real-time traffic data, such as video feeds and sensory inputs, could improve the MLLMs' understanding of dynamic traffic situations. Finally, enhancing the explainability of MLLMs in traffic behavior scenarios will help users understand the rationale behind model predictions.}




\section{Access to the Benchmark and Datasets}

\subsection{Availability}
The Traffic Behavior Benchmark (TB-Bench) and the training datasets (TB-100k, TB-250k) will be publicly available at the following Github repository: 
\begin{itemize}
    \item \url{https://github.com/TB-AD/TB-Bench-110k-250k} 
\end{itemize}

\qt{The source code for conducting and analyzing the experiments will also be publicly available in the repository upon publication, permitting free use for research purposes.}

\vspace{-0.3cm}



\qt{
\subsection{Future Update}
We also plan to establish an evaluation server and leaderboard on HuggingFace in the future. 
Any updates will be communicated through the above Github repository to ensure users have access to the latest information.
}

\section{Benchmark and Datasets}

\subsection{Task Definition}

\subsubsection{Relative Distance (RD).} 
\qt{The task is to predict the Euclidean distance in meters between two entities in an image; see Figure \ref{fig:benchmark_task1_example} for two examples.}

\subsubsection{Spatial Reasoning (SR).} 
\qt{The task is to predict the spatial position of one entity relative to another from the perspective of a reference entity; see Figure \ref{fig:benchmark_task2_example} for examples. Specifically, the relationship between two objects is defined by the angle \(\theta\), as follows:}

\begin{equation}
\text{Relation} = 
\begin{cases} 
\text{front} & \text{if } -30^\circ < \theta \leq 30^\circ, \\
\text{front left} & \text{if } 30^\circ < \theta \leq 90^\circ, \\
\text{front right} & \text{if } -90^\circ < \theta \leq -30^\circ, \\
\text{back left} & \text{if } 90^\circ < \theta \leq 150^\circ, \\
\text{back right} & \text{if } -150^\circ < \theta \leq -90^\circ, \\
\text{back} & \text{otherwise}.
\end{cases}
\end{equation}

This angular relationship is similar to that defined in \cite{nuscene-qa}.


\subsubsection{Orientation Reasoning (OR).} 
\qt{This task is to predict the facing relationship between two entities from the perspective of a reference entity, categorized as: `similar', `opposite', or `perpendicular'. Please refer to Figure \ref{fig:benchmark_task3_example} for examples. The relationship is defined based on the absolute difference in facing angles \(|\theta|\), as follows:}

\begin{equation}
\text{Relation} = 
\begin{cases} 
\text{similar} & \text{if } 0^\circ \leq |\theta| \leq 45^\circ, \\
\text{opposite} & \text{if } 135^\circ \leq |\theta| \leq 180^\circ, \\
\text{perpendicular} & \text{otherwise.}
\end{cases}
\end{equation}

\qt{It is noted that this angle is measured from the facing direction of a reference entity to the position of the target entity in Euclidean space, irrespective of the target entity's facing direction.}


\subsubsection{Other Lane to Ego-Vehicle (EGO-LANE).} \qt{This task is to predict the lane of a target vehicle relative to the ego-vehicle's perspective; see Figure \ref{fig:benchmark_task4_example} for examples. The categories include: `front lane', `front left lane', `front right lane', and `oncoming traffic lane' (the lane on the opposite side of the road).}

\qt{It is noted that when the ego-vehicle is on a road with multiple lanes, the `front lane' is further classified into three fine-grained categories: `front lane', `front left lane', and `front right lane'.}


\subsubsection{Other Lane Changing (OBJ-LANE).} 
\qt{ This task is to predict whether the target vehicle is changing lanes, categorized as `left lane change', `right lane change', or `no change'; see Figure \ref{fig:benchmark_task5_example} for examples. Lane changes are evaluated based on the target vehicle's viewpoint. For instance, if the target vehicle in the oncoming traffic lane executes a right lane change, the ego vehicle perceives it as moving to the left.}


\subsubsection{Other Turning (OBJ-TURN).} \qt{This task is to predict whether the target vehicle is making a turn, categorized as `turning left', `turning right', or `go straight'. The target vehicle is considered to be turning, if it changes direction by more than 25 degrees within a period of 1.6 seconds. Please refer to Figure \ref{fig:benchmark_task6_example} for examples.}


\subsubsection{Ego Turning (EGO-TURN).} \qt{This task is to predict whether the ego-vehicle is making a turn, categorized as turning left, turning right, or going straight. The turning maneuver of the ego-vehicle is also defined by a change in direction of more than 25 degrees within a period of 1.6 seconds. Please refer to Figure \ref{fig:benchmark_task7_example} for examples.}


\subsubsection{Ego Traverse Distance (EGO-TRA).} \qt{This task is to predict the traverse distance of the ego vehicle in meters over a period of 1.6 seconds. Please see Figure \ref{fig:benchmark_task8_example} for examples.}


\subsection{Dataset Statistics} \label{data_stat}



\qt{Table \ref{tab:stats_tb_bench}, \ref{tab:stats_tb_100k}, and \ref{tab:stats_tb_250k} show the distribution of categories for the TB-Bench, TB-100k, and TB-250k datasets, respectively, detailing the count and percentage of samples for various task types.}

\qt{To create the TB-Bench, we manually screened the frames thoroughly to select samples with clearly visible target entities. Each task in TB-Bench has an equal count of 250 samples. We ensure that the distribution of categories in each task closely resembles that of the instruction tuning datasets.}

\qt{It is seen from Table \ref{tab:stats_tb_100k}, and \ref{tab:stats_tb_250k} that TB-250k represents a normal scene occurrence distribution in real-world scenarios, while TB-100k is a more label-balanced version.}

 
\begin{table}[h!]
    \caption{\textbf{TB-Bench Statistics}}
    \centering
    \renewcommand{\arraystretch}{1.20} 
    \resizebox{0.48\textwidth}{!}{
    \begin{tabular}{
        |>{\centering\arraybackslash}p{0.3\linewidth}|
        >{\centering\arraybackslash}p{0.36\linewidth}|
        >{\centering\arraybackslash}p{0.12\linewidth}|
        >{\centering\arraybackslash}p{0.3\linewidth}|
        }
        \hline
        \textbf{Task Type} & \textbf{Category} & \textbf{Count} & \textbf{Percentage (\%)} \\
        \hline
        Relative Distance & numerical value & 250 & 12.5 \\
        \hline
         & back & 61 & 3.0 \\
         & back left & 30 & 1.5 \\
        Spatial  & back right & 9 & 0.4 \\
        Reasoning & front & 87 & 4.3 \\
         & front left & 45 & 2.2 \\
         & front right & 18 & 0.9 \\
        \hline
        & numerical value & 122 & 6.1 \\
        Orientation  & opposite & 51 & 2.5 \\
        Reasoning & perpendicular & 16 & 0.8 \\
         & similar & 61 & 3.0 \\
        \hline
        & front lane & 71 & 3.5 \\
        Other Lane to & front left lane & 40 & 2.0 \\
        Ego-Vehicle & front right lane & 31 & 1.6 \\
         & oncoming traffic lane & 108 & 5.4 \\
        \hline
        Other Lane  & left lane change & 62 & 3.1 \\
        Changing & no change & 142 & 7.1 \\
         & right lane change & 46 & 2.3 \\
        \hline
        & go straight & 126 & 6.3 \\
        Other Turning & left turn & 67 & 3.4 \\
         & right turn & 57 & 2.9 \\
        \hline
        & go straight & 122 & 6.1 \\
        Ego Turning & left turn & 38 & 1.9 \\
         & right turn & 90 & 4.5 \\
        \hline
        Ego Traverse Distance & numerical value & 250 & 12.5 \\
        \hline

    \end{tabular}
    }
    \label{tab:stats_tb_bench}
\end{table}

\begin{table}[h!]
    \caption{\textbf{TB-100k Statistics}}
    \centering
    \renewcommand{\arraystretch}{1.20}
    \resizebox{0.48\textwidth}{!}{
    \begin{tabular}{
        |>{\centering\arraybackslash}p{0.3\linewidth}|
        >{\centering\arraybackslash}p{0.36\linewidth}|
        >{\centering\arraybackslash}p{0.12\linewidth}|
        >{\centering\arraybackslash}p{0.30\linewidth}|
        }
        \hline
        \textbf{Task Type} & \textbf{Category} & \textbf{Count} & \textbf{Percentage (\%)} \\
        \hline
        Relative Distance & numerical value & 10000 & 9.1 \\
        \hline
        \multirow{6}{*}{Spatial Reasoning} & back & 3580 & 3.3 \\
         & back left & 3183 & 2.9 \\
         & back right & 3115 & 2.8 \\
         & front & 7873 & 7.2 \\
         & front left & 7321 & 6.7 \\
         & front right & 4928 & 4.5 \\
        \hline
        & numerical value & 10000 & 9.1 \\
         Orientation & opposite & 10013 & 9.1 \\
         Reasoning & perpendicular & 2387 & 2.2 \\
         & similar & 7600 & 6.9 \\
        \hline
        & front lane & 3889 & 3.5 \\
        Other Lane to & front left lane & 3231 & 2.9 \\
        Ego-Vehicle & front right lane & 4182 & 3.8 \\
         & oncoming traffic lane & 8698 & 7.9 \\
        \hline
        Other Lane  & left lane change & 414 & 0.4 \\
        Changing & no change & 807 & 0.7 \\
         & right lane change & 279 & 0.3 \\
        \hline
        & go straight & 744 & 0.7 \\
        Other Turning & left turn & 435 & 0.4 \\
         & right turn & 321 & 0.3 \\
        \hline
        \multirow{3}{*}{Ego Turning} & go straight & 753 & 0.7 \\
         & left turn & 331 & 0.3 \\
         & right turn & 416 & 0.4 \\
        \hline
        Ego Traverse Distance & numerical value & 15500 & 14.1 \\
        \hline
    \end{tabular}
    }
    \label{tab:stats_tb_100k}
\end{table}

\begin{table}[h!]
    \caption{\textbf{TB-250k Statistics}}
    \centering
    \renewcommand{\arraystretch}{1.20}
    \resizebox{0.48\textwidth}{!}{
    \begin{tabular}{
        |>{\centering\arraybackslash}p{0.3\linewidth}|
        >{\centering\arraybackslash}p{0.36\linewidth}|
        >{\centering\arraybackslash}p{0.12\linewidth}|
        >{\centering\arraybackslash}p{0.30\linewidth}|
        }
        \hline
        \textbf{Task Type} & \textbf{Category} & \textbf{Count} & \textbf{Percentage (\%)} \\
        \hline
        Relative Distance & numerical value & 34721 & 13.7 \\
        \hline
        \multirow{6}{*}{Spatial Reasoning} & back & 17023 & 6.7 \\
         & back left & 6247 & 2.5 \\
         & back right & 3966 & 1.6 \\
         & front & 26917 & 10.6 \\
         & front left & 10793 & 4.3 \\
         & front right & 4804 & 1.9 \\
        \hline
        & numerical value & 34872 & 13.7 \\
        Orientation & opposite & 19242 & 7.6 \\
        Reasoning & perpendicular & 3355 & 1.3 \\
         & similar & 12283 & 4.8 \\
        \hline
        & front lane & 14312 & 5.6 \\
        Other Lane to & front left lane & 4454 & 1.8 \\
        Ego-Vehicle & front right lane & 6401 & 2.5 \\
         & oncoming traffic lane & 24833 & 9.8 \\
        \hline
        Other Lane  & left lane change & 414 & 0.2 \\
        Changing & no change & 807 & 0.3 \\
         & right lane change & 279 & 0.1 \\
        \hline
        \multirow{3}{*}{Other Turning} & go straight & 744 & 0.3 \\
         & left turn & 435 & 0.2 \\
         & right turn & 321 & 0.1 \\
        \hline
        \multirow{3}{*}{Ego Turning} & go straight & 753 & 0.3 \\
         & left turn & 331 & 0.1 \\
         & right turn & 416 & 0.2 \\
        \hline
        Ego Traverse Distance & numerical value & 25000 & 9.9 \\
        \hline
    \end{tabular}
    }
    \label{tab:stats_tb_250k}
\end{table}

\section{Data Generation Pipeline Details}


\subsection{Information Extraction} 

\begin{figure*}[t]
\centering
\includegraphics[width=0.99\linewidth]{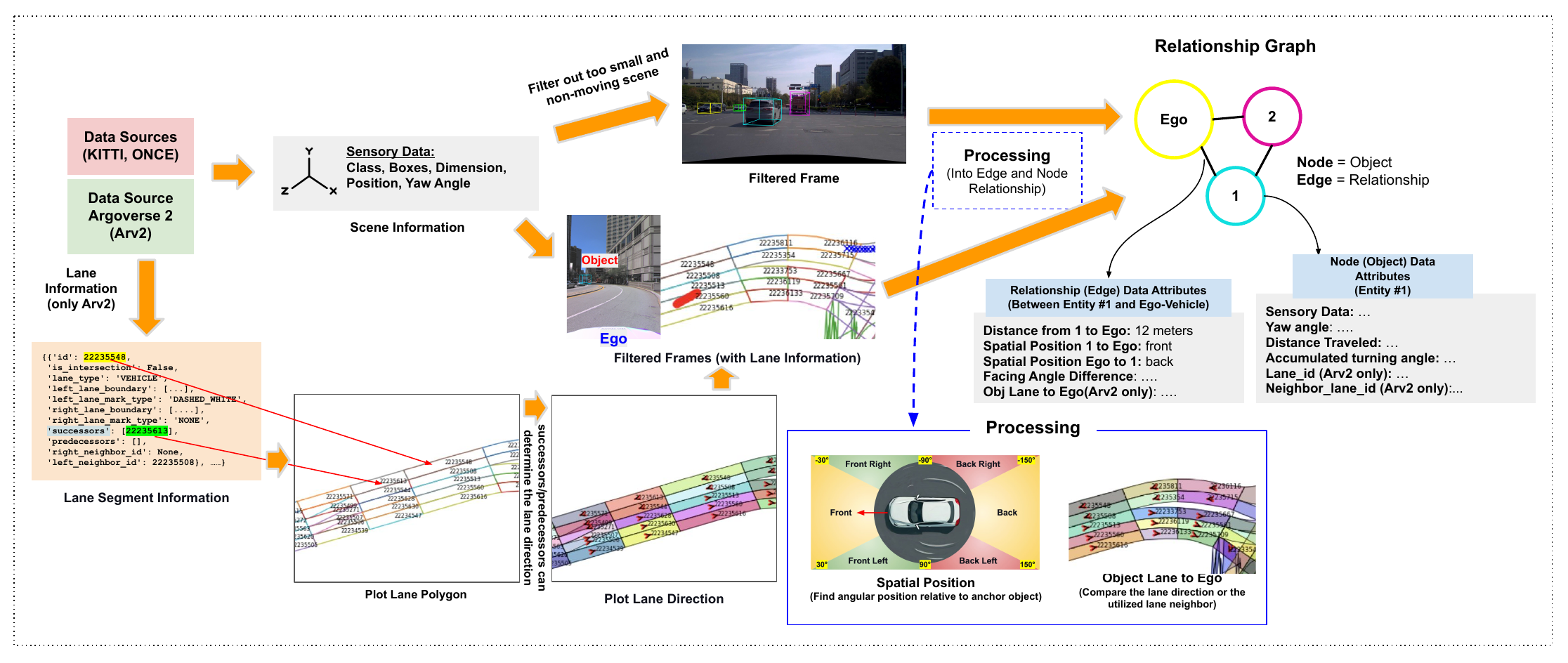}
\caption{\textbf{Data Extraction Process.}}
\label{fig:additional_data_extraction}
\end{figure*}

\qt{Figure \ref{fig:additional_data_extraction} shows the extraction process. It begins with obtaining raw sensory data from input samples, which may include static images with entity attributes from datasets like KITTI or ONCE, or sequential data from Argoverse2. This sensory data is processed to filter out insignificant scene information.}


\qt{For Argoverse2, lane geometry information is processed concurrently. Lane coordinates are used to create polygons with attributes, such as neighboring, successor, and predecessor lanes. This information helps determine lane direction and angle, which are then projected onto vehicle attributes to obtain the vehicle's lane ID and relevant lane information. This data is subsequently passed to the next processing step to extract all scene attributes.}

\begin{table*}[t!]
    \caption{\textbf{Q\&A Templates.} 
    \qt{The placeholder \texttt{<entity\_n>} refers to any entity, such as `Entity \#1', `Entity \#2', or `Ego-vehicle', ensuring that no sentence contains duplicate entities. `Short Answer Template' denotes a basic class of concise responses that can be expanded into more complex sentences.}
    } 
    \centering
    \renewcommand{\arraystretch}{1.2}
    \resizebox{0.98\linewidth}{!}{
    \begin{tabular}{
    |>{\centering\arraybackslash}p{0.12\linewidth}|>{\arraybackslash}p{0.70\linewidth}|>
    {\arraybackslash}p{0.25\linewidth}|
    }
        \hline
        \textbf{Task Type} & \textbf{Question Template} & \textbf{Short Answer Template} \\
        \hline
        Relative Distance & Can you measure straight-line distance in meters between \texttt{<entity\_n>} and \texttt{<entity\_n>}? & xx.xx meters \\
         & How far is \texttt{<entity\_n>} from \texttt{<entity\_n>} in meters? & \\
         & How many meters apart are \texttt{<entity\_n>} and \texttt{<entity\_n>}? & \\
        & What is distance from \texttt{<entity\_n>} to \texttt{<entity\_n>} along road's surface in meters? & \\
        \hline
        Spatial Reasoning & How are \texttt{<entity\_n>} and \texttt{<entity\_n>} spatially related, from \texttt{<entity\_n>} perspective? & back, back left, back right, front, front left, front right \\
         & What is spatial position of \texttt{<entity\_n>} relative to \texttt{<entity\_n>}? & \\
        & What is spatial relation of \texttt{<entity\_n>} to \texttt{<entity\_n>}? & \\
        \hline
        Orientation Reasoning & How do you describe orientation of \texttt{<entity\_n>} relative to \texttt{<entity\_n>}, similar, opposite or perpendicular? & opposite, perpendicular, similar, xx.xx degrees \\
        & How is \texttt{<entity\_n>} oriented relative to \texttt{<entity\_n>}, similar, opposite or perpendicular? & \\
         & What is angle between \texttt{<entity\_n>} and \texttt{<entity\_n>}, in degrees? & \\
         & What is facing angle of \texttt{<entity\_n>} relative to \texttt{<entity\_n>}, in degrees? & \\
        & What is orientation of \texttt{<entity\_n>} relative to \texttt{<entity\_n>}, similar, opposite or perpendicular? & \\
        & What is yaw angle different between \texttt{<entity\_n>} and \texttt{<entity\_n>}, in degrees? & \\
        \hline
        Other Lane to Ego-Vehicle & How would you describe lane position of Entity\#1? Options: front lane, front left lane, front right lane, or oncoming traffic lane. & front\_lane, front\_left\_lane, front\_right\_lane, oncoming\_traffic\_lane \\
         & & \\
        \hline
        Other Lane Changing & How would you describe driving scene involving Entity\#1? Please explain, focusing on vehicle's lane change maneuver. & left\_lane\_change, no\_change, right\_lane\_change \\
         & & \\
        \hline
        Other Turning & How would you describe driving scene involving Entity\#1? Please explain, focusing on vehicle's turning maneuver. & go\_straight, left\_turn, right\_turn \\
        \hline
        Ego Turning & How would you describe driving scene involving our car? Please explain, focusing on our car's turning maneuver. & go\_straight, left\_turn, right\_turn \\
        \hline
        Ego Traverse Distance & How far has our car driven and what kind of steering maneuver did it perform in current scene? & xx.xx meters \\
         & & \\
        \hline
    \end{tabular}
    }
    \label{tab:question_answer_templates}
\end{table*}

\subsection{Rule-based Q\&A Generation}


\qt{The process begins with obtaining attribute data from either the nodes or edges of the relationship graph. This data is then processed through rule-based functions to extract behavioral or spatial information. Next, we generate behavioral attributes in a Q\&A format using templates provided in Table \ref{tab:question_answer_templates}.}

\qt{Generation depends on the task type. Tasks 1-4 and task 8 (`Relative Distance,' `Spatial Reasoning,' `Orientation Reasoning,' `Other Lane to Ego,' and `Ego Traverse Distance') can be created in any frame, as their attributes are available in all frames.}

\qt{In contrast, tasks 5-7 (`Other Lane Changing,' Other Turning,' and Ego Turning') require a triggering event, specifically a change in attributes. The following details explain how to trigger an event:}


\begin{tcolorbox}[title=Event Triggering: Other Lane Changing]
    \begin{itemize}
        \item Check if the current \texttt{lane\_id} is in the \texttt{future\_right\_neighbor\_id}. \\ 
        If yes, then assign: \textbf{Right Lane Change}.
        \item Check if the current \texttt{lane\_id} is in the \texttt{future\_left\_neighbor\_id}. \\ 
        If yes, then assign: \textbf{Left Lane Change}.
        \item If neither condition is met, 
        assign: \textbf{No Change}.
    \end{itemize}
    Note: \texttt{future\_right\_neighbor\_id} refers to the \texttt{right\_neighbor\_id} of the next time step; the same applies to the left side.
\end{tcolorbox}

\begin{tcolorbox}[title=\qt{Event Triggering:} Other Turning]
    \begin{itemize}
        \item Check if the accumulated object yaw angle is greater than 25 degrees in 1.6 seconds. \\ If yes, then assign: \textbf{Turn Left}.
        \item Check if the accumulated object yaw angle is less than -25 degrees in 1.6 seconds. \\ If yes, then assign: \textbf{Turn Right}.
        \item If neither condition is met, assign: \textbf{Go straight}.
    \end{itemize}
\end{tcolorbox}

\begin{tcolorbox}[title=\qt{Event Triggering:} Ego Turning]
    \begin{itemize}
        \item Check if the accumulated ego-vehicle yaw angle is greater than 25 degrees in 1.6 seconds. \\ If yes, then assign: \textbf{Turn Left}.
        \item Check if the accumulated ego-vehicle yaw angle is less than -25 degrees in 1.6 seconds. \\ If yes, then assign: \textbf{Turn Right}.
        \item If neither condition is met, assign: \textbf{Go straight}.
    \end{itemize}
\end{tcolorbox}

\subsection{Q\&A Augmentation}


\qt{The augmentation process converts short question-answer (Q\&A) pairs into natural language sentences. Each short QA pair was expanded into a full sentence using a predefined structure. We employ the Microsoft-Phi3-medium model to generate these sentences, using the following prompt:}

\begin{tcolorbox}[title=Complete Prompt]
\footnotesize
\texttt{system\_text = "You are a language expert assistant. In this task, we want to expand the following answer to longer wording but no additional information."}

\texttt{\textbf{full\_prompt} = f"\{system\_text\}. The question is: \{question\} and the short answer is \{answer\}. Give the complex answer in a short sentence no more than 15 words."}
\end{tcolorbox}

\qt{The parameters for \texttt{\{question\}}, and \texttt{\{answer\}} are dynamically inserted for each instance. This approach ensures that the augmented data remains concise (up to 15 words) while incorporating the original short answer in a more elaborated context, maintaining the correctness and relevance of the response.}





\subsection{Pre-crash Scenarios}
\begin{figure}[h!]
\centering
\includegraphics[width=1.0\linewidth]{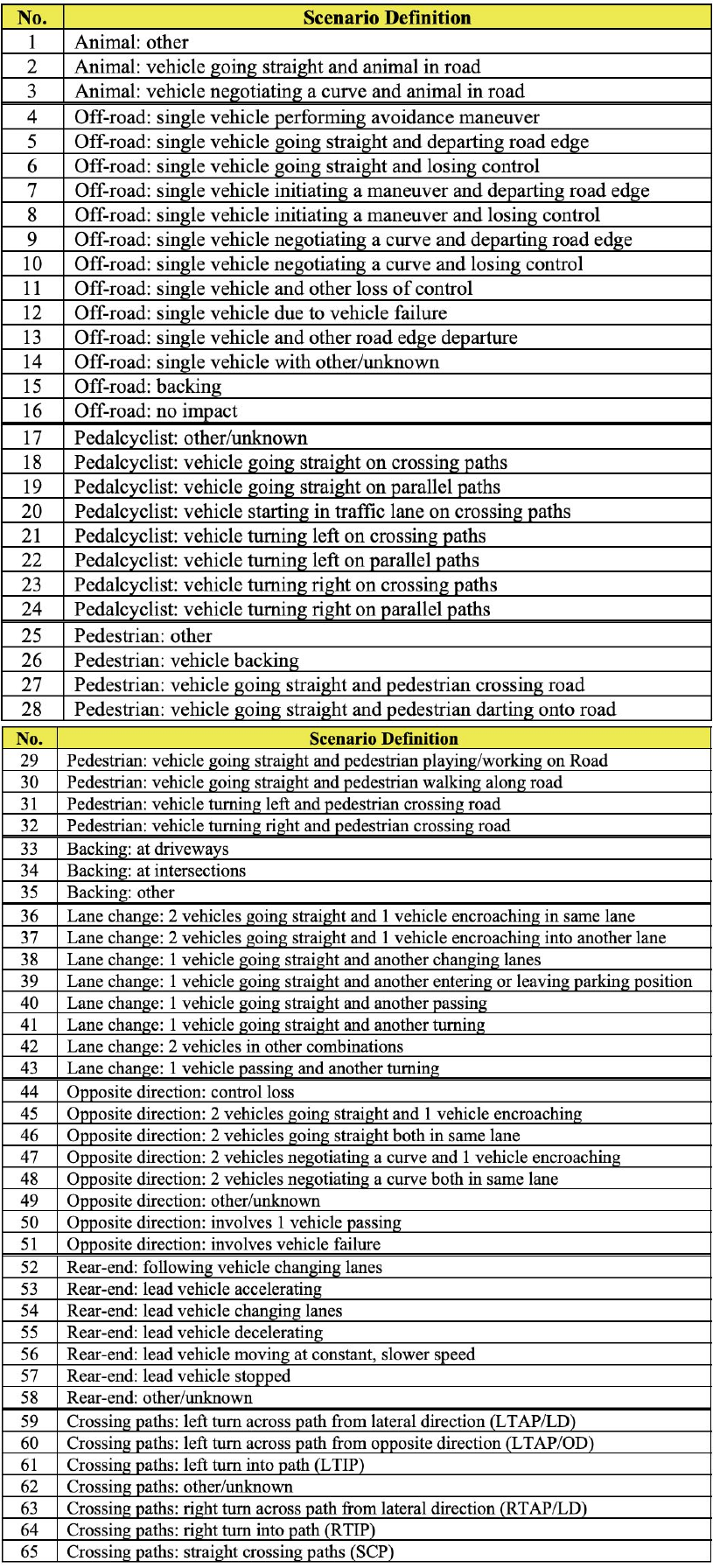}
\caption{\qt{List of pre-crash scenarios based on National Automotive Sampling System (NASS) variables.}}
\label{fig:pre_crash_scenarios}
\end{figure}

\qt{Figure ~\ref{fig:pre_crash_scenarios} presents the full list of 65 pre-crash scenarios as described in Section Task Design, based on National Automotive Sampling System. Each scenario is categorized into a specific accident type, such as `Animal', `Off-road', etc.} 

\section{Evaluation Details}

\subsection{Evaluation Metrics}

\begin{figure*}[h!]
\centering
\includegraphics[width=0.8\linewidth]{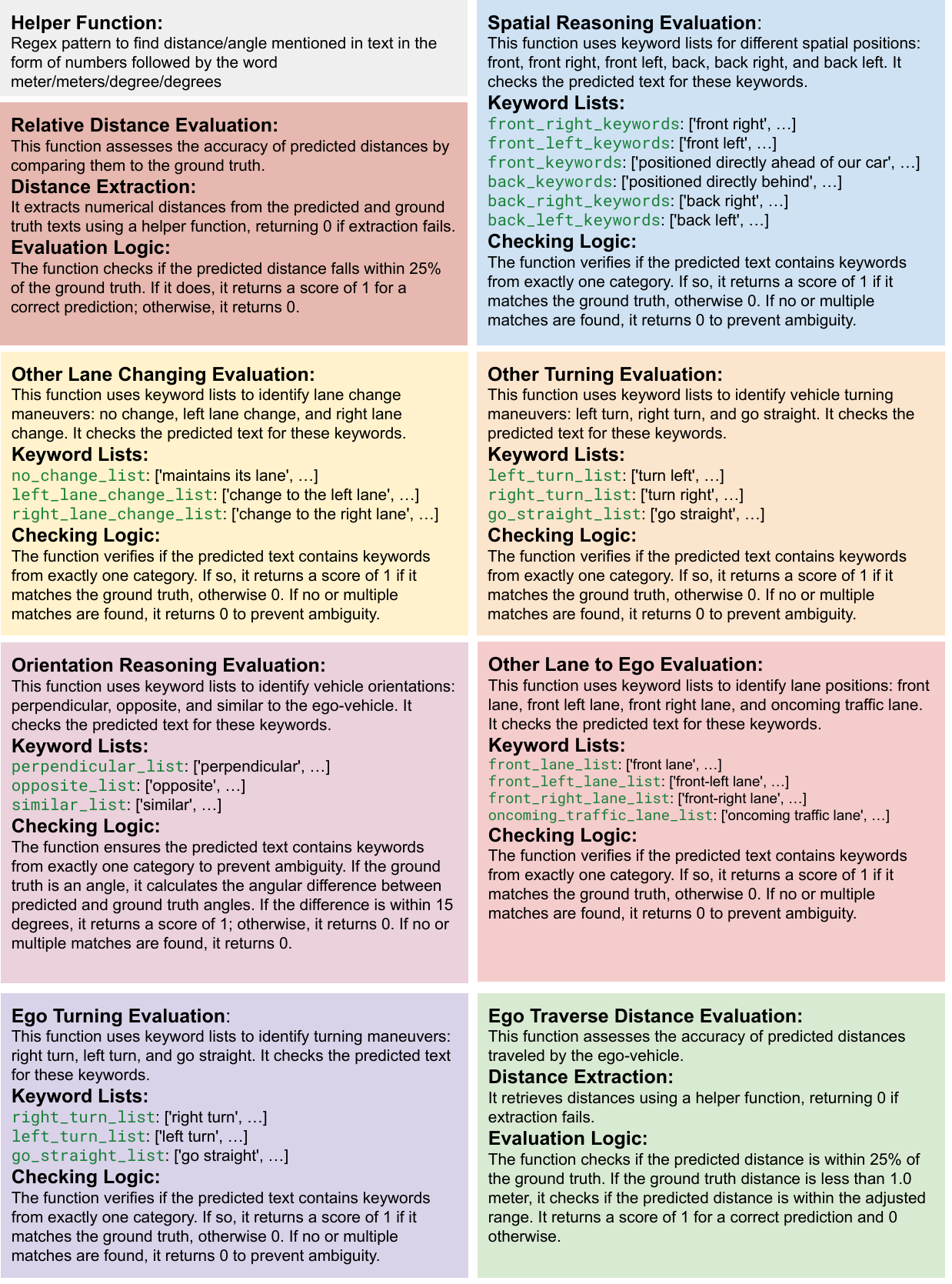}
\caption{\textbf{Evaluation Metric Methodology for Each Task:} The method uses rule-based and regular expressions techniques to assess accuracy.}
\label{fig:evaluation_metric_function}
\end{figure*}

\qt{As mentioned in the main paper, we employ the rule-based methods for evaluation. Figure \ref{fig:evaluation_metric_function} shows the keyword list and regular expression used in the evaluation pipeline}.



\subsection{Additional Details on Evaluated Models}
\qt{In this study, we evaluate open-source state-of-the-art models and proprietary models on our TB-Bench in a zero-shot manner. We provide additional information for the evaluated models in Table \ref{tab:huggingface_name_api}.}

\qt{The first category consists of open-source models (LLaVA, Bunny, and InternVL), which are accessible via the Hugging Face API. These models are fully fine-tuned with specific settings for each version available in their Huggingface repositories.}

\qt{The second category consists of proprietary models (GPT-4o and Gemini), which require specific API calls and image formatting. It is noted that we evaluate the latest version of these models on our TB-Bench at the time of submission.} 


\begin{table*}[ht]
    \caption{\qt{\textbf{Addtional information of the models evaluated on TB-Bench.}}
    }
    \centering
    \resizebox{1.0\textwidth}{!}{
    \begin{tabular}{
        >{\arraybackslash}p{0.3\linewidth}
        >{\arraybackslash}p{0.45\linewidth}
        >{\centering\arraybackslash}p{0.22\linewidth}
        >{\centering\arraybackslash}p{0.22\linewidth}
        }
        \toprule
        \textbf{Model Name} & \textbf{Full Repository/API Name} & \textbf{Vision Part} & \textbf{Language Part} \\
        \midrule
        \rowcolor{gray!10}\textbf{Open-source models} & & & \\

        LLaVA-1.5-7B & llava-hf/llava-1.5-7b-hf & CLIP-L/14 & Vicuna-7b-v1.5 \\
        LLaVA-v1.6-Mistral-7B & llava-hf/llava-v1.6-mistral-7b-hf & CLIP-L/14 & Mistral-7B-Instruct-v0.2  \\
        LLaVA-NeXT-Video-7B & llava-hf/LLaVA-NeXT-Video-7B-hf & CLIP-L/14 & Vicuna-7B-v1.5 \\
        LLaVA-Interleave-Qwen-7B & llava-hf/llava-interleave-qwen-7b-hf & SigLIP-L/14 & Qwen1.5-7B-Chat \\
        Bunny-v1.1-4B & BAAI/Bunny-v1\_1-4B & SigLIP-L/14 & Phi-3-mini-4k-instruct \\
        Bunny-v1.1-Llama-3-8B-V & BAAI/Bunny-v1\_1-Llama-3-8B-V & SigLIP-L/14 & Llama-3-8B-Instruct \\
        InternVL2-8B & OpenGVLab/InternVL2-8B & InternViT-300M-448px & Qwen2-8B-Instruct \\
        Mini-InternVL2-1B-DriveLM & OpenGVLab/Mini-InternVL2-1B-DA-DriveLM & InternViT-300M-448px & Qwen2-0.5B\\
        DriveLM-mantis-8b & francepfl/DriveLM-mantis-8b-idefics2\_8192 &  SiGLIP & Mistral-7B-v0.1 \\
        \rowcolor{gray!10}\textbf{Proprietary models} & & & \\
        Gemini-1.5-flash & Gemini-1.5-flash & Unknown & Unknown \\
        GPT-4o-2024-08-06 & GPT-4o-2024-08-06 & Unknown & Unknown \\
        \bottomrule
    \end{tabular}
    }
    \label{tab:huggingface_name_api}
\end{table*}

\subsection{Prompt for Zero-Shot Evaluation}
\qt{For zero-shot evaluation of existing models, we use an Option Template that presents multiple-choice options to define possible answer classes. This approach accommodates the varied terminology that pre-trained models may employ to describe situations.}

\qt{The details of the Option Template, which varies based on the task type, are as follows:}

\begin{tcolorbox}[title=Option Template]
\footnotesize
\textbf{Distance-Related Tasks:}
\begin{itemize}
    \item \texttt{Answer in xx.x meters format.}
\end{itemize}

\textbf{Angle-Related Tasks:}
\begin{itemize}
    \item \texttt{Answer in xx.x degrees format.}
\end{itemize}

\textbf{Tasks with Predefined Answer Choices:}
\begin{itemize}
    \item Retrieve the answer choices.
    \item Assign a letter to each choice (e.g., A, B, C).
    \item Present options as follows:
    \begin{quote}
        \texttt{Options: \\ A. choice1, \\ B. choice2, \\ C. choice3, ...}
    \end{quote}
\end{itemize}
\end{tcolorbox}

\qt{Pre-trained models often use specific vocabularies based on their training data. For instance, a model might say `opposite side of the road' instead of `oncoming traffic lane' if it lacks specific instruction training. By offering explicit choices, the model can select the appropriate terminology despite variations.}

\qt{For numerical answers, we specify the expected format within the prompt to ensure clarity and consistency, such as instructing the model to \texttt{Answer in xx.x meters format.}}

\qt{This structured approach allows the model to account for variations in wording and select the most appropriate option, demonstrating its understanding.}

\section{Experiments and Results}

\subsection{Implementation Details}
\begin{table}[ht!]
    \caption{\qt{Hyper-parameter settings for finetuning our models on TB-100k or TB-250k.}} 
    \centering
    \begin{tabular}{
            >{\arraybackslash}p{0.2\textwidth}
            >{\centering\arraybackslash}p{0.15\textwidth}
            }
        \toprule
            \textbf{Hyper-parameter} & \textbf{Value} \\
        \midrule
            Epochs & {10} \\
            Warmup steps & {2,000} \\
            Learning rate & {1e-5} \\
            LoRA learning rate & {1e-4} \\
            Effective Batch size & {64} \\
            AdamW $\beta$ & {(0.9, 0.999)} \\
            Weight decay & {0.05} \\
            Drop path & {0} \\
            Attention dropout & {0} \\
            Torch data type & bf16 \\
            Inference temperature & {0} \\

        \bottomrule
    \end{tabular}
    \label{tab:hyperparameters}
\end{table}

All models are finetuned on an Ubuntu 20.04 server equipped with four A6000 GPUs, each with 48GB of memory. The source code is built on the Transformers library \cite{wolf2019huggingface} and utilizes the PyTorch 2.4 framework \cite{paszke2019pytorch}.

\qt{Additional information on hyper-parameter settings for finetuning our baseline models on TB-100k and TB-250k is presented in Table \ref{tab:hyperparameters}.}


\subsection{Quantitative Analyses}
\qt{We provide quantitative analyses and the qualitative results of the model's predictions on TB-Bench. The baseline model ((SigLIP-L/14 and Qwen1.5-0.5b) finetuned on TB-100k.}
\qt{For numerical output tasks, we visualize error distributions using box plots. On the other hand, we use confusion matrices for classification tasks.}


\subsubsection{Relative Distance and Ego Traverse Distance Tasks.}

\begin{figure}[h!]
\centering
\includegraphics[width=0.998\linewidth]{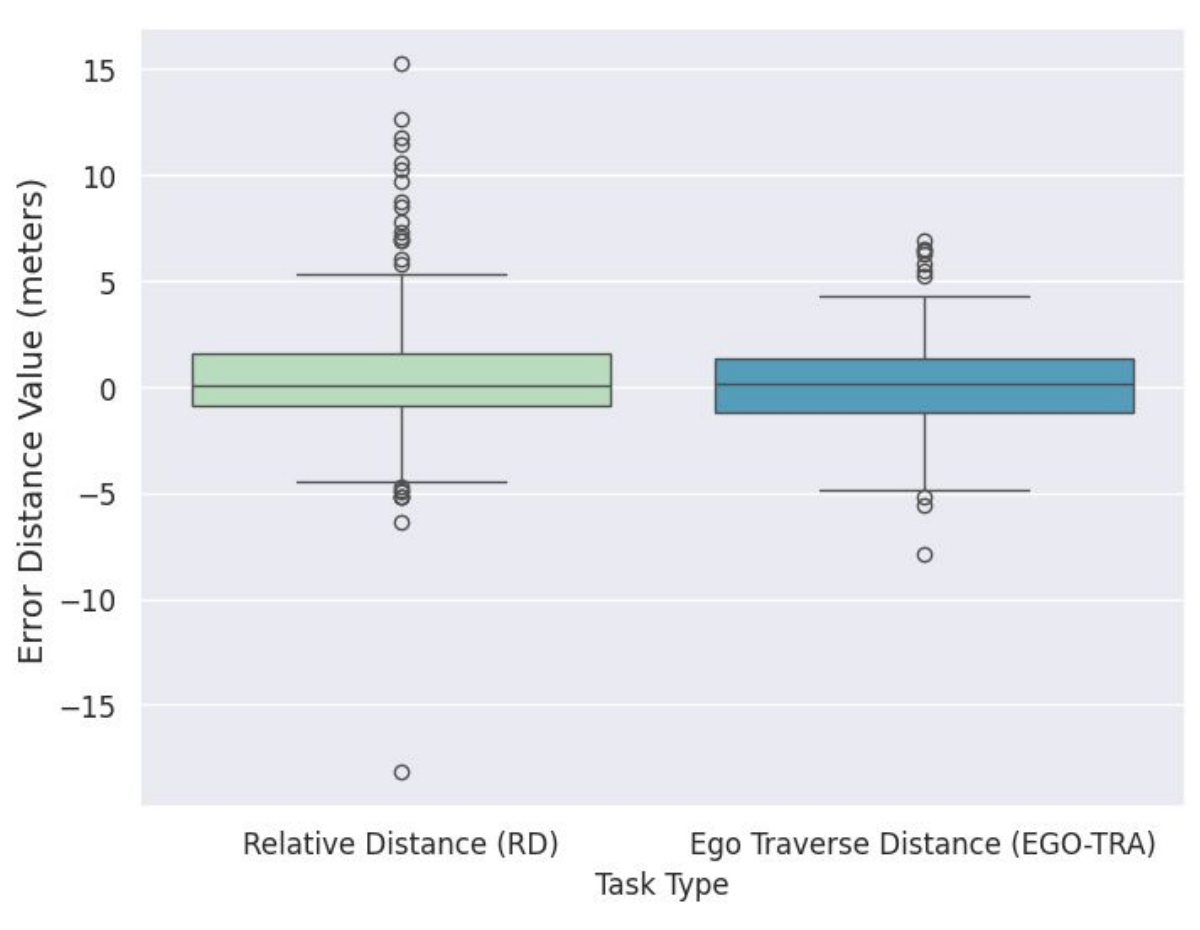}
\caption{
\qt{Distance error on Relative Distance (RD) and Ego Traverse Distance (EGO-TRA) tasks.}
}
\label{fig:quant_score_task1_8}
\end{figure}

\qt{Figure \ref{fig:quant_score_task1_8} shows the box plot for distance errors of our model predictions on the two tasks. For RD, distance errors are generally centered around zero, with a narrow interquartile range, indicating consistent performance, though a few outliers suggest overestimation. Predictions on EGO-TRA show a similar error distribution, with the median slightly above zero and more positive outliers, indicating a tendency to overestimate distance.}

\subsubsection{Orientation Reasoning Task.}

\begin{figure}[h!]
\centering
\includegraphics[width=0.998\linewidth]{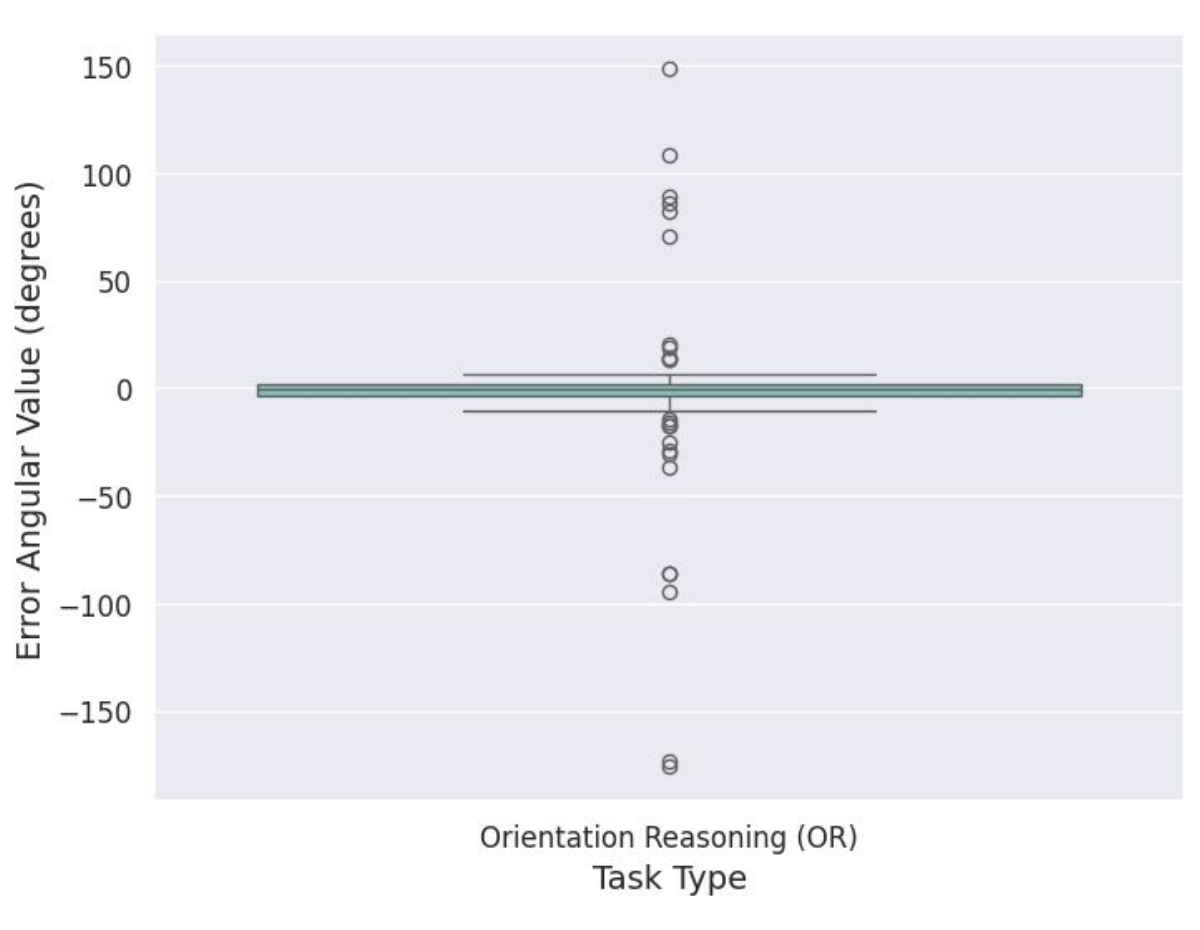}
\caption{
\qt{Angular error on Orientation Reasoning (OR) task.}
}
\label{fig:quant_score_task3}
\end{figure}
\qt{Figure \ref{fig:quant_score_task3} shows the box plot for angular errors of our model predictions on the Orientation Reasoning (OR) task. The median and interquartile range are close to zero, indicating precise and consistent predictions. Short whiskers further highlight this accuracy. Outliers are grouped near 0, 90, and 180 degrees, suggesting small angle misestimations. Overall, the model demonstrates minimal errors in this task.}


\begin{figure}[t!]
\centering
\includegraphics[width=0.998\linewidth]{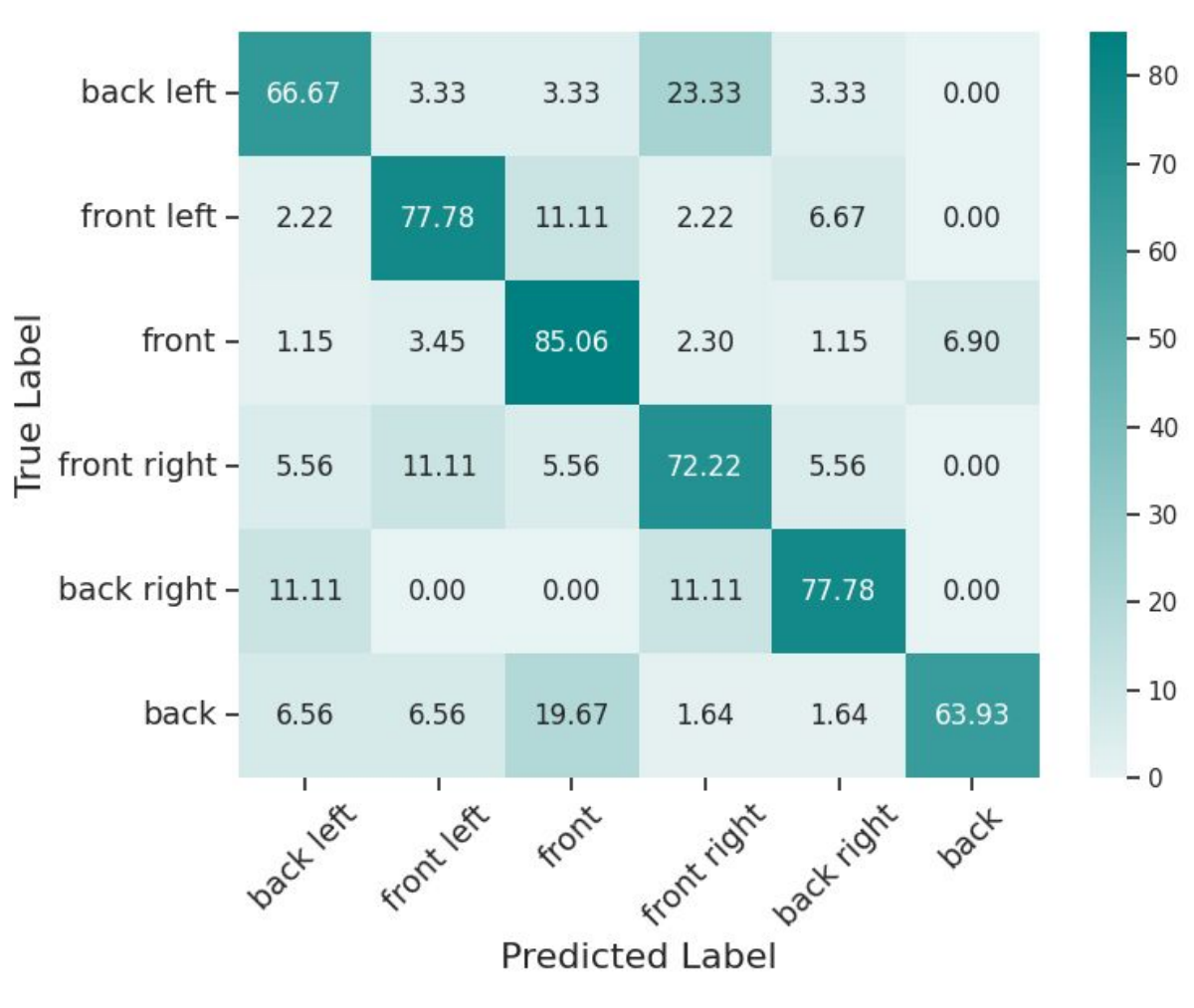}
\caption{\qt{Confusion matrix on Spatial Reasoning (SR) task.}}
\label{fig:quant_class_task2}
\end{figure}

\subsubsection{Spatial Reasoning Task.}
\qt{Figure \ref{fig:quant_class_task2}
shows the confusion matrix of our model predictions on the Spatial Reasoning (SR) task. The `front' position is classified most accurately at 85.1\%, while `back' and `back left' positions have lower accuracies of 63.3\% and 66.7\%. The matrix also shows moderate confusion between similar positions, such as back left' being misclassified as front right' (23.33\%) and `back' as `front' (19.67\%).}



\subsubsection{Other Lane to Ego-Vehicle Task.}
\begin{figure}[t]
\centering
\includegraphics[width=0.998\linewidth]{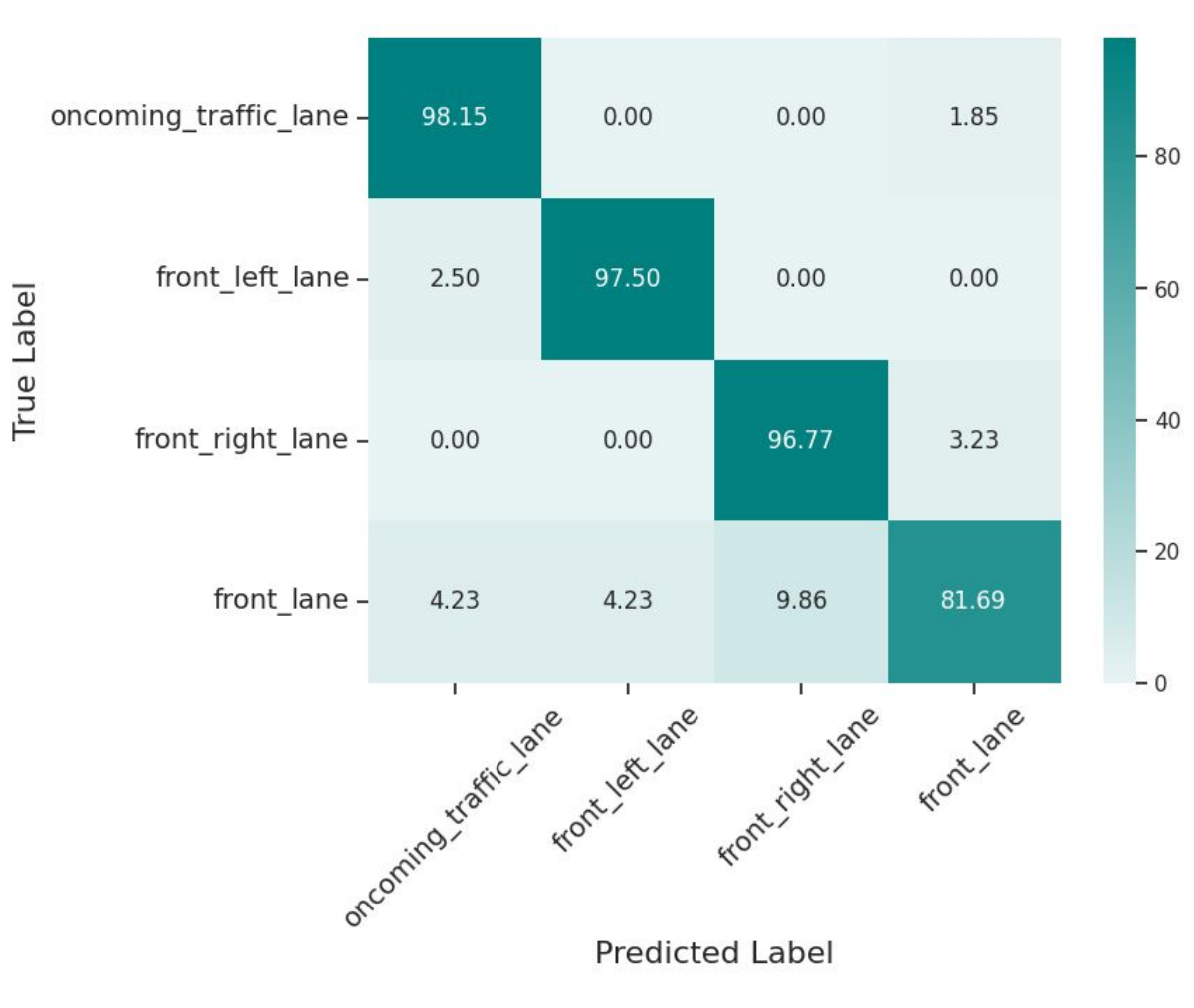}
\caption{\qt{Confusion Matrix on Other Lane to Ego-Vehicle (EGO-LANE).}}
\label{fig:quant_class_task4}
\end{figure}

\qt{Figure \ref{fig:quant_class_task4} shows the confusion matrix of our model predictions on the Other Lane to Ego-Vehicle (EGO-LANE) task. Overall, the model shows high accuracy on most categories (over 96\%), except for the `front lane,' which has an accuracy of only 81.7\%. The primary misclassification pattern involves confusion between the `front lane' and its adjacent lanes, with 9.9\% of `front lane' samples being misclassified as `front right lane.'}

\subsubsection{Other Lane Changing Task.}
\begin{figure}[ht]
\centering
\includegraphics[width=0.998\linewidth]{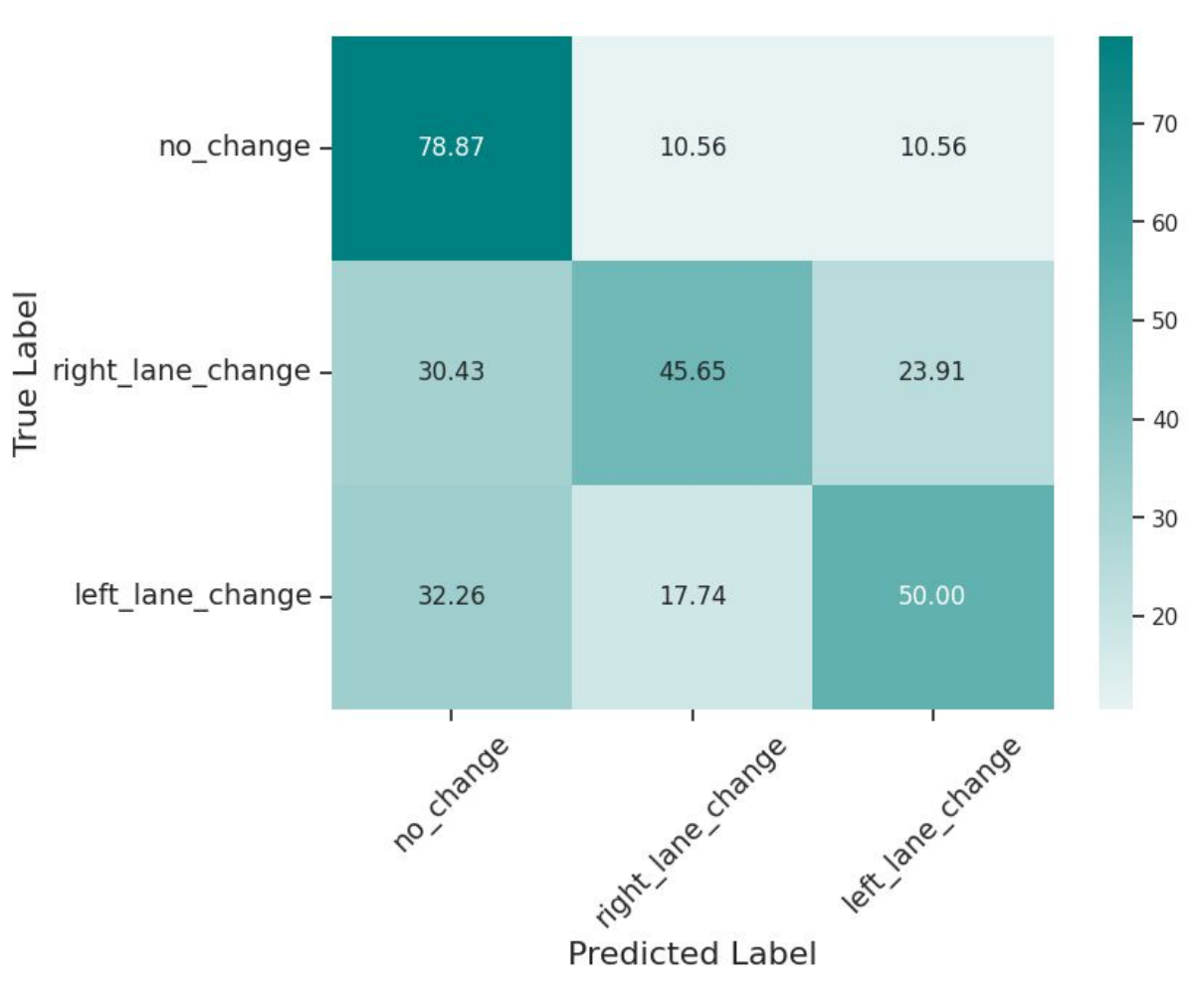}
\caption{\qt{Confusion Matrix on Other Lane Changing (OBJ-LANE).}}
\label{fig:quant_class_task5}
\end{figure}

\qt{Figure \ref{fig:quant_class_task5} shows the confusion matrix on the Other Lane Changing (OBJ-LANE) task, where samples are categorized into `no change,' `left lane change,' and `right lane change.' In this case, the model shows decent performance with an accuracy of around 78.87\% in the `no change' category. However, it struggles significantly with lane change predictions. For both `left lane change' and `right lane change' classifications, the most misclassified predictions are in the `no change' category, with 32.3\% and 30.4\% misclassified, respectively. This indicates the model's difficulty in distinguishing between lane changes and no change, underscoring the task's challenges.}

\subsubsection{Other Lane Changing Task.}
\begin{figure}[ht]
\centering
\includegraphics[width=0.998\linewidth]{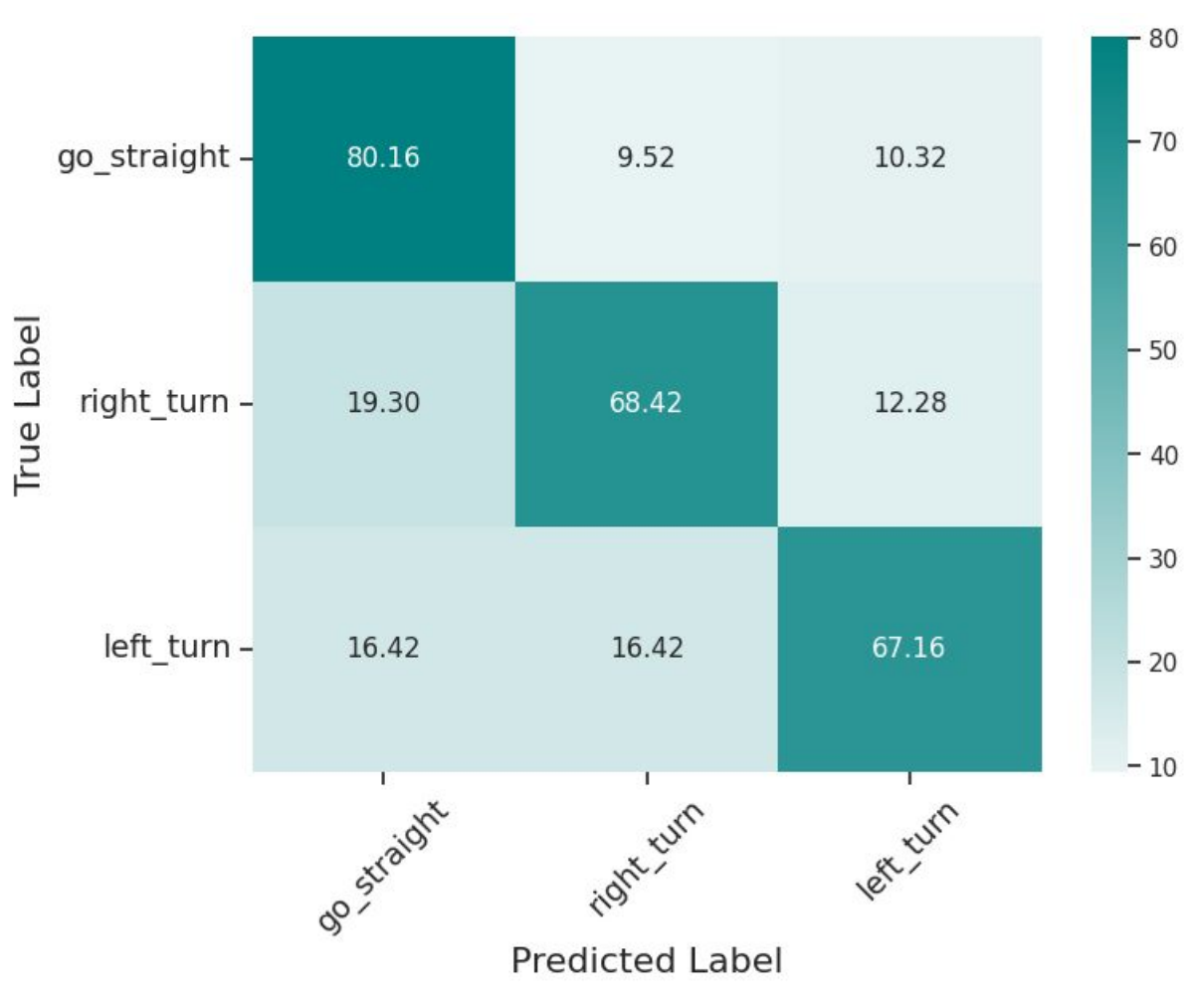}
\caption{\qt{Confusion matrix on Other Turning (OBJ-TURN).}}
\label{fig:quant_class_task6}
\end{figure}

\qt{Figure \ref{fig:quant_class_task6} shows the confusion matrix on the Other Turning (OBJ-TURN) task, where samples are categorized as `left turn,' `go straight,' and `right turn.' The model excels in identifying the go straight' category, achieving an accuracy of 80.16\%. However, it shows over 30\% misclassification rates for both `left turn' and `right turn.' Notably, misclassifications of `left turn' are nearly evenly divided between `right turn' and go `straight,' despite `right turn' errors being more theoretically opposed. The model's performance indicates that it struggles to accurately interpret turns from the perspective of other vehicles, influenced by road orientation and vehicle positioning.
}

\subsubsection{Ego Turning Task.}

\begin{figure}[t!]
\centering
\includegraphics[width=0.998\linewidth]{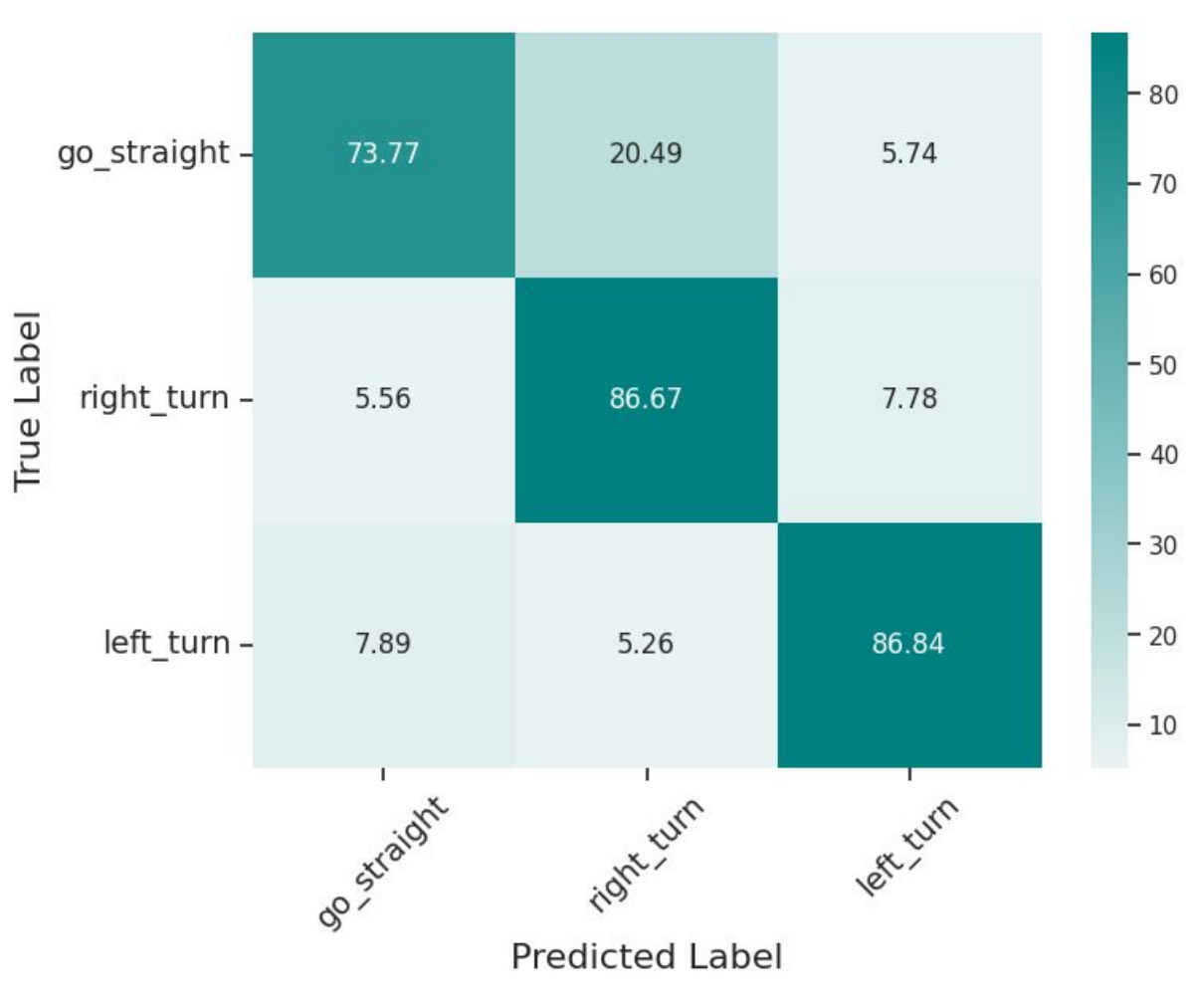}
\caption{\qt{Confusion matrix on Ego Turning (EGO-TURN)}}
\label{fig:quant_class_task7}
\end{figure}
\qt{Figure \ref{fig:quant_class_task7} shows the confusion matrix on the task, where the actions are categorized as `left turn,' `go straight,' and `right turn.' The model demonstrates strong performance in identifying turns, with high accuracy rates of 86.8\% for `left turn' and 86.67\% for `right turn.' Interestingly, the turning maneuvers have stronger performance than the `go straight' action, with a notable 20.49\% of `go straight' samples being misclassified as `right turn.' 
}



\subsection{\qt{Qualitative Results}}
\qt{For brevity, we present two samples per task, each with input frame(s), the task question, and the ground truth answer. Each sample also includes predictions from our fine-tuned baseline model (SigLIP-L/14 and Qwen1.5-0.5b) and the best performing zero-shot model, GPT-4o (GPT-4o-2024-08-06 version).}

\qt{Figures for each task are as follows:}

\begin{figure*}[t]
\centering
\includegraphics[width=0.998\linewidth]{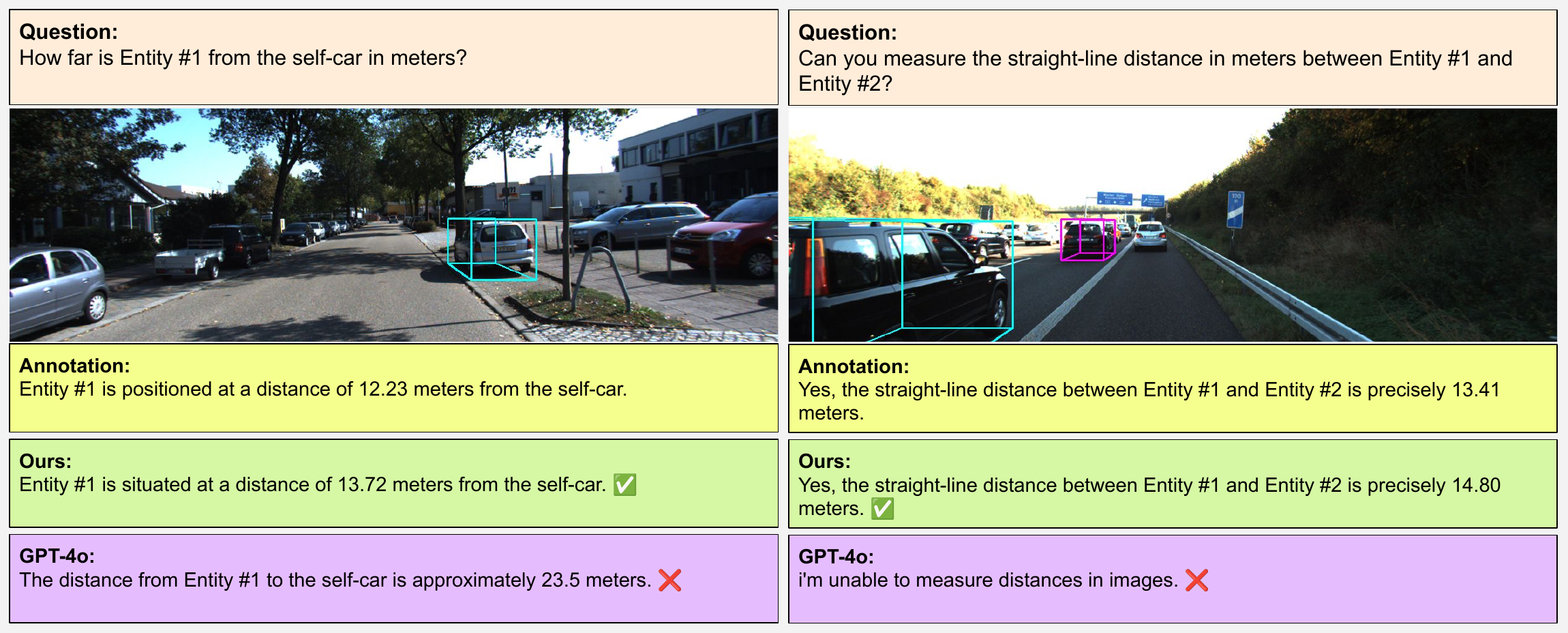}
\caption{\qt{Examples and predictions from our baseline method and GPT-4o for the Relative Distance (RD) task.}}
\label{fig:benchmark_task1_example}
\end{figure*}

\begin{figure*}[t]
\centering
\includegraphics[width=0.998\linewidth]{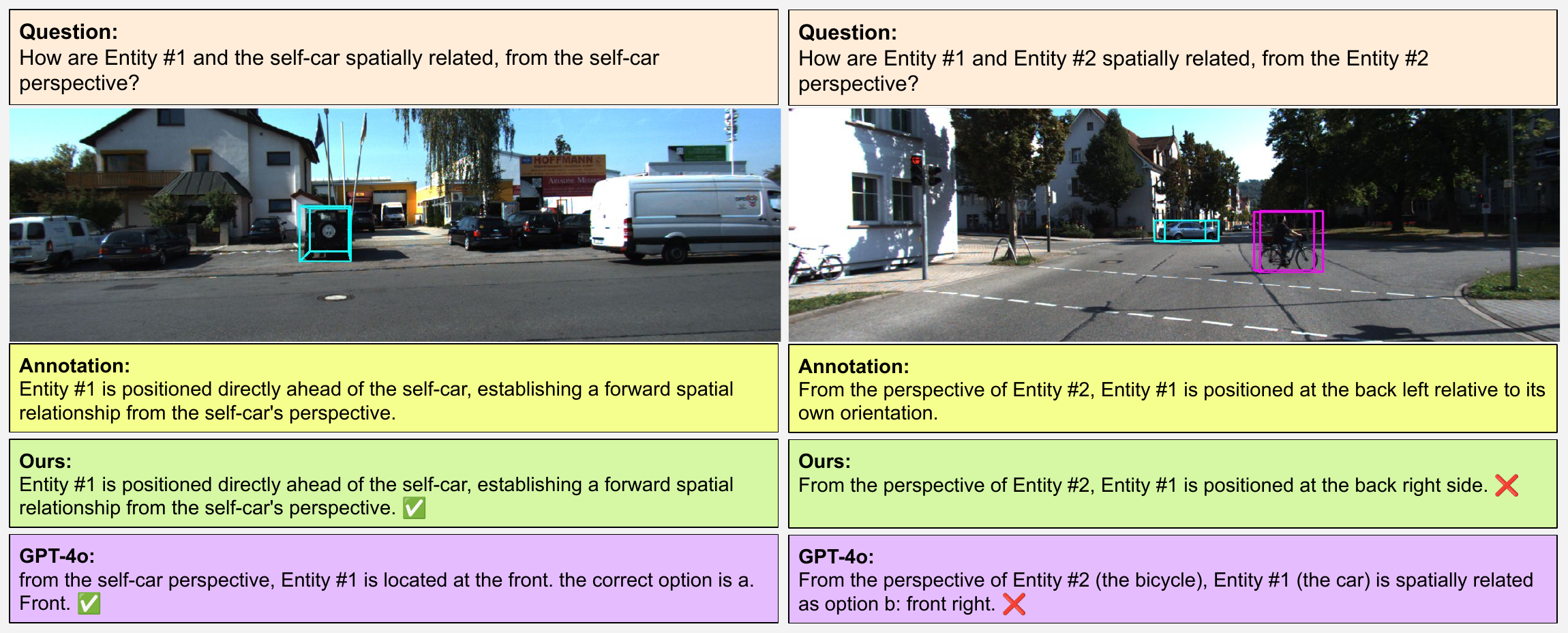}
\caption{\qt{Examples and predictions from our baseline method and GPT-4o for the Spatial Reasoning (SR) task.}}
\label{fig:benchmark_task2_example}
\end{figure*}

\begin{figure*}[t]
\centering
\includegraphics[width=0.998\linewidth]{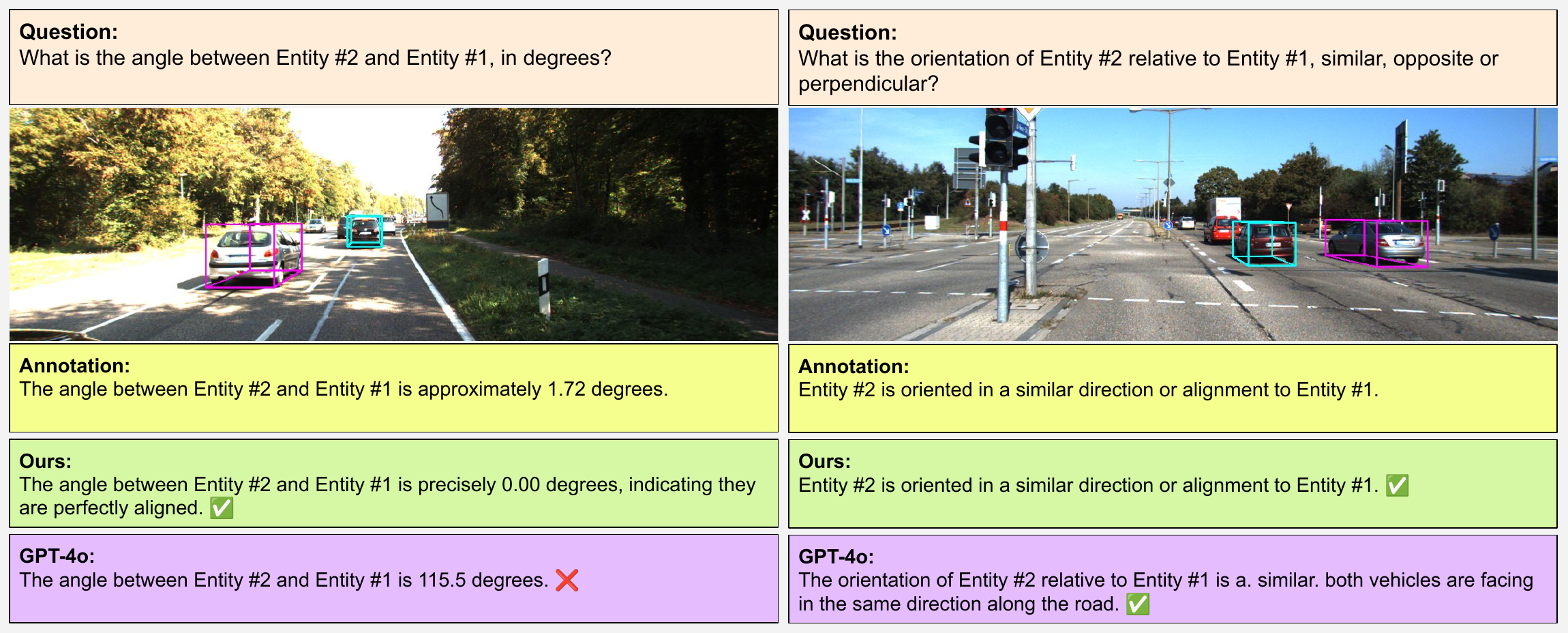}
\caption{\qt{Examples and predictions from our baseline method and GPT-4o for the Orientation Reasoning (OR) task.}}
\label{fig:benchmark_task3_example}
\end{figure*}

\begin{figure*}[t]
\centering
\includegraphics[width=0.998\linewidth]{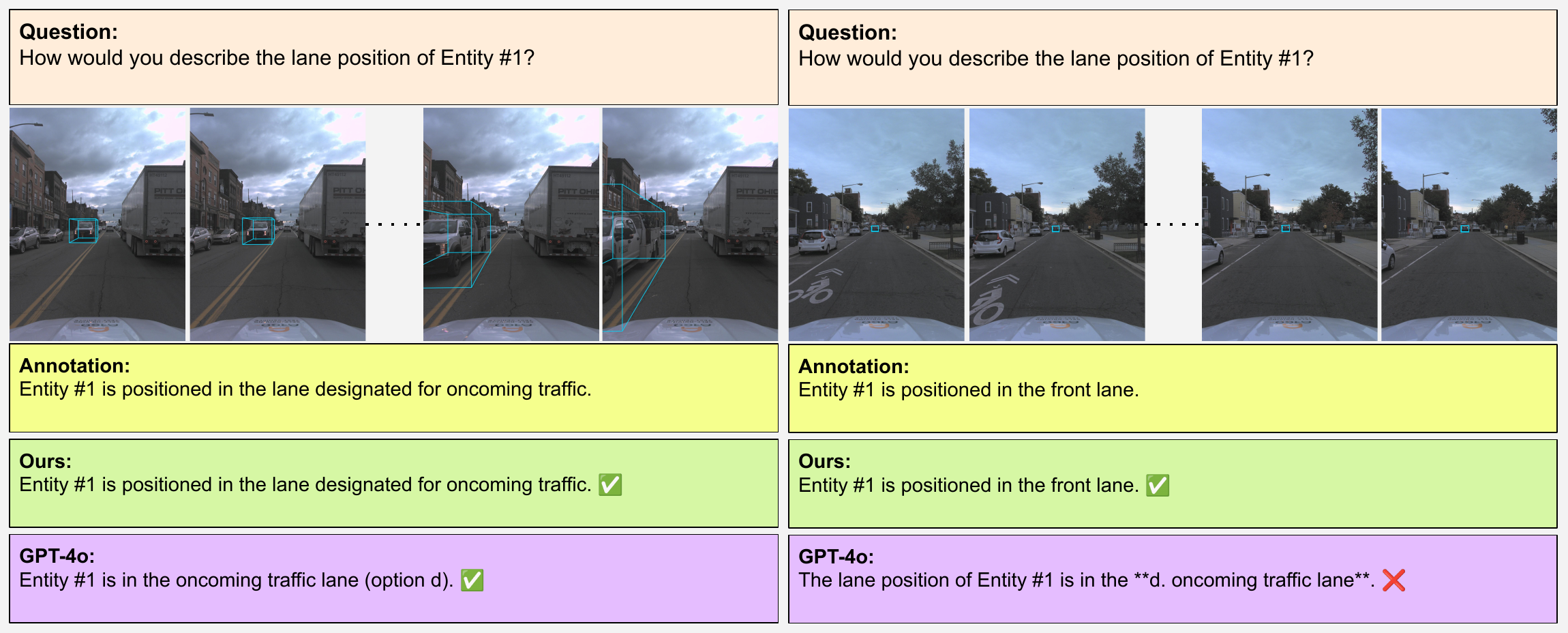}
\caption{\qt{Examples and predictions from our baseline method and GPT-4o for the Other Lane to Ego-Vehicle (EGO-LANE) task.}}
\label{fig:benchmark_task4_example}
\end{figure*}

\begin{figure*}[t]
\centering
\includegraphics[width=0.998\linewidth]{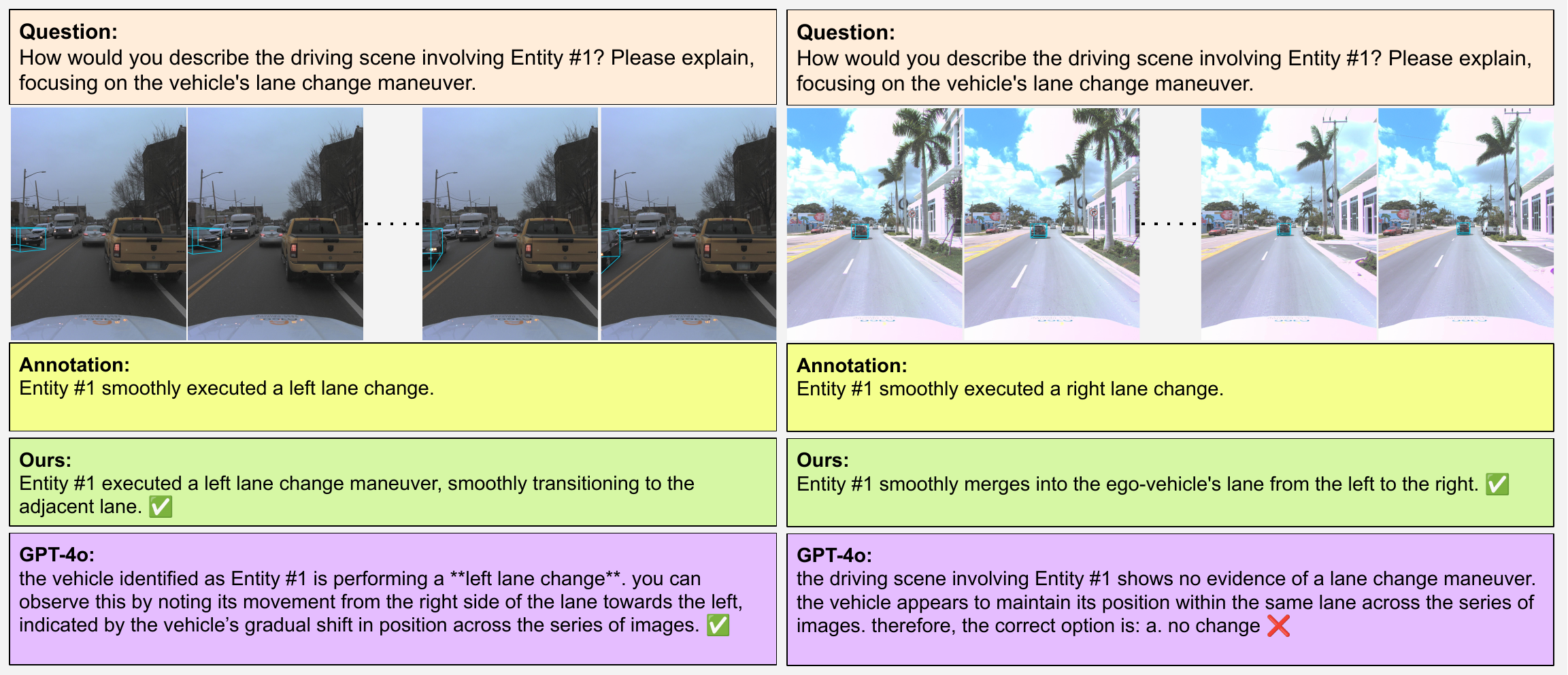}
\caption{\qt{Examples and predictions from our baseline method and GPT-4o for the Other Lane Changing (OBJ-LANE) task.}}
\label{fig:benchmark_task5_example}
\end{figure*}

\begin{figure*}[t]
\centering
\includegraphics[width=0.998\linewidth]{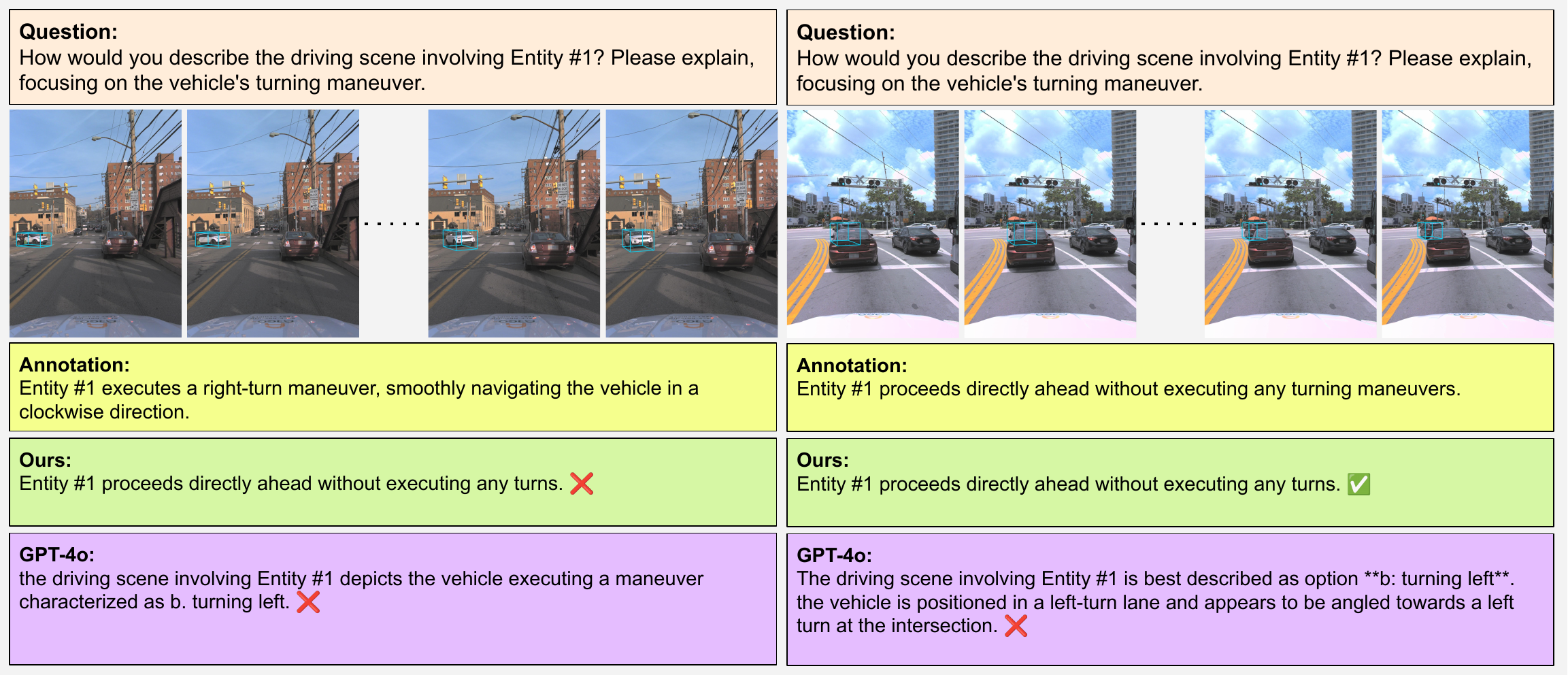}
\caption{\qt{Examples and predictions from our baseline method and GPT-4o for the Other Turning (OBJ-TURN) task.}}
\label{fig:benchmark_task6_example}
\end{figure*}

\begin{figure*}[t]
\centering
\includegraphics[width=0.998\linewidth]{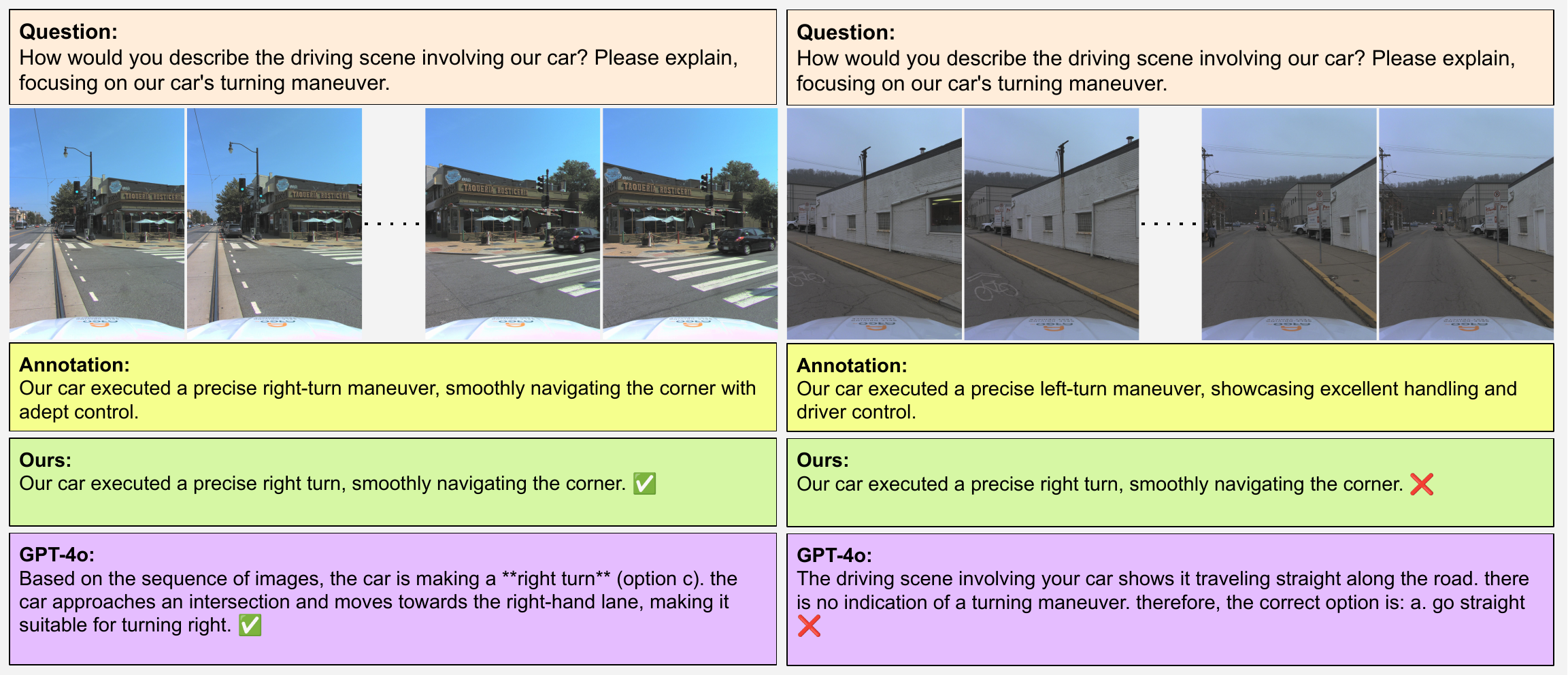}
\caption{\qt{Examples and predictions from our baseline method and GPT-4o for the Ego Turning (EGO-TURN) task.}}
\label{fig:benchmark_task7_example}
\end{figure*}

\begin{figure*}[t]
\centering
\includegraphics[width=0.998\linewidth]{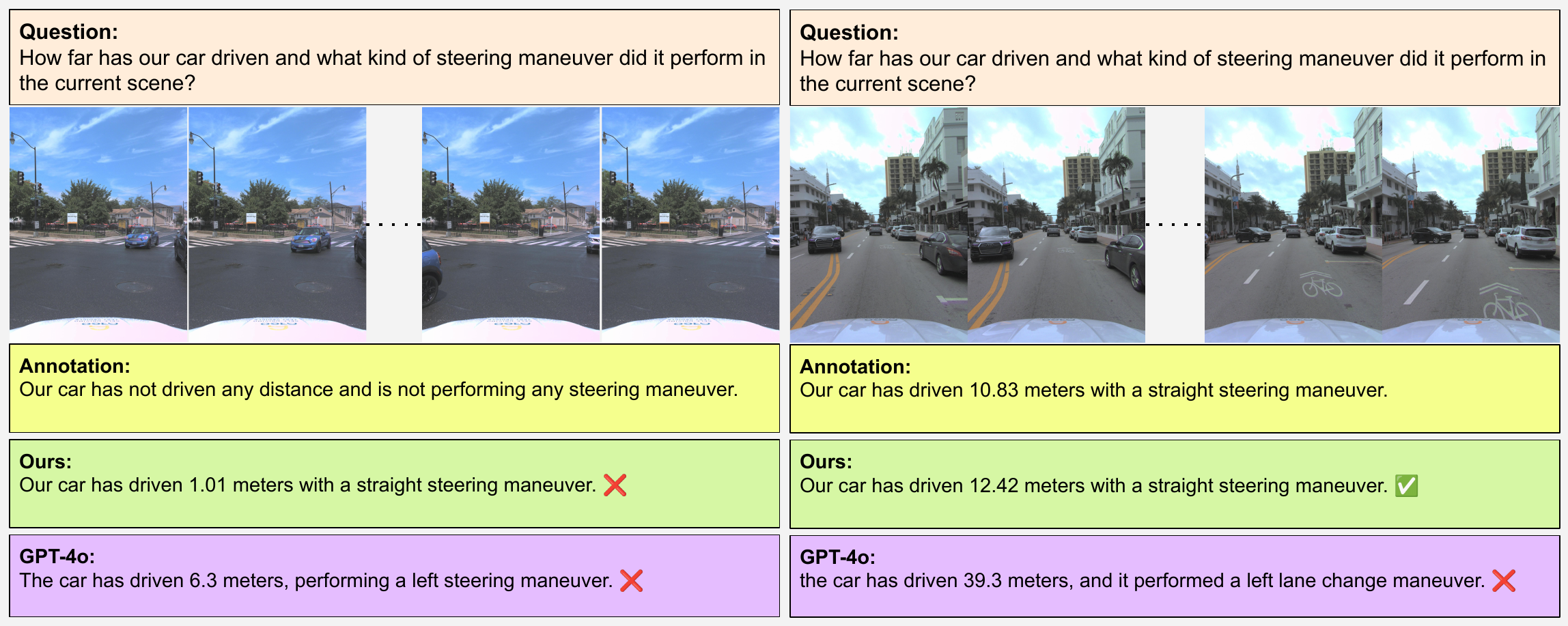}
\caption{\qt{Examples and predictions from our baseline method and GPT-4o for the Ego Traverse Distance (EGO-TRA) task.}}
\label{fig:benchmark_task8_example}
\end{figure*}

\begin{itemize}
    \item Figure \ref{fig:benchmark_task1_example}: Relative Distance (RD)

    \item Figure \ref{fig:benchmark_task2_example}: Spatial Reasoning (SR)

    \item Figure \ref{fig:benchmark_task3_example}: Orientation Reasoning (OR)

    \item Figure \ref{fig:benchmark_task4_example}: Other Lane to Ego-Vehicle (EGO-LANE)

    \item Figure \ref{fig:benchmark_task5_example}: Other Lane Changing (OBJ-LANE)

    \item Figure \ref{fig:benchmark_task6_example}: Other Turning (OBJ-TURN) task

    \item Figure \ref{fig:benchmark_task7_example}: Ego Turning (EGO-TURN) 

    \item Figure \ref{fig:benchmark_task8_example}: Ego Traverse Distance (EGO-TRA) 
\end{itemize}

\subsection{Ablation Study Details}

\qt{We provide detailed ablation results across eight tasks in Table \ref{tab:ablation_result}.}



\qt{Results indicate that stronger visual encoders significantly improve performance. For instance, comparing CLIP-L/14 to SigLIP-L/14 shows improvements of over 15.2\% in Relative Distance (RD), 4.0\% in Orientation Reasoning (OR), 5.6\% in Other Turning (OBJ-TURN), and 10.4\% in Ego Turning (EGO-TURN).}


\qt{The optimal number of visual tokens is 16. Increasing this to 36 tokens improves Ego Traverse Distance (EGO-TRA) by only 2.8\%, while performance in other tasks declines compared to the 16-token variant.}


\qt{Utilizing more sequential frames generally enhances performance, especially in the tasks requiring temporal information (tasks 3-8). Single-frame tasks like Spatial Reasoning also benefit from training on multi-frame tasks, showing notable improvements. For ego-focused tasks, using 8 frames instead of 2 results in significant gains of over 14\% in EGO-TURN and 12.8\% in EGO-TRA, indicating that the number of frames is more critical for ego-focused tasks than for object-focused ones.}

\begin{table*}[t]
\centering
\caption{
\qt{\textbf{Ablation results per task}. All the models are finetuned on the TB-100k dataset, with their performance evaluated on TB-Bench and reported in accuracy (percentage).}
}
\resizebox{0.98\textwidth}{!}{%
\begin{tabular}
    {
    >{\arraybackslash}p{0.26\textwidth}
    >{\centering\arraybackslash}p{0.06\textwidth}
    >{\centering\arraybackslash}p{0.06\textwidth}
    >{\centering\arraybackslash}p{0.06\textwidth}
    >{\centering\arraybackslash}p{0.11\textwidth}
    >{\centering\arraybackslash}p{0.15\textwidth}
    >{\centering\arraybackslash}p{0.11\textwidth}
    >{\centering\arraybackslash}p{0.11\textwidth}
    >{\centering\arraybackslash}p{0.11\textwidth}
    >{\centering\arraybackslash}p{0.06\textwidth}
    }
\toprule
    \multicolumn{1}{c}{\multirow{3}{*}{\Large\textbf{Model}}} & 
    \multicolumn{9}{c}{\multirow{2}{*}{\Large\textbf{TrafficBehaviorBenchmark (TB-Bench)}}}
    \\
    \\
    & \multirow{1}{*}{\textbf{RD} $\uparrow$} 
    & \multirow{1}{*}{\textbf{SR} $\uparrow$} 
    & \multirow{1}{*}{\textbf{OR} $\uparrow$} 
    & \multirow{1}{*}{\textbf{EGO-LANE} $\uparrow$} 
    & \multirow{1}{*}{\textbf{OBJ-LANE} $\uparrow$}
    & \multirow{1}{*}{\textbf{OBJ-TURN} $\uparrow$}
    & \multirow{1}{*}{\textbf{EGO-TURN} $\uparrow$}
    & \multirow{1}{*}{\textbf{EGO-TRA} $\uparrow$}
    & \multirow{1}{*}{\textbf{Avg} $\uparrow$}
    \\
\midrule
    \midrule
    \rowcolor{gray!10}\textbf{Visual encoder} & & & & & & & & & \\
    CLIP-L/14 & 61.2 & 72.8 & 82.8 & 91.6 & 61.2 & 69.2 & 70.8 & 66.0 & 72.0 \\
    SigLIP-B/16 & 65.2 & 70.4 & 86.8 & 90.4 & 70.0 & 69.6 & 75.2 & 65.6 & 74.3 \\
    SigLIP-L/14 & 76.4 & 74.4 & 86.8 & 94.0 & 68.8 & 74.8 & 81.2 & 63.2 & 77.5 \\
    \midrule
    \rowcolor{gray!10}\textbf{Visual tokens per frame} & & & & & & & & & \\
    4 & 68.8 & 70.0 & 86.4 & 94.0 & 67.6 & 74.0 & 71.6 & 49.2 & 72.7 \\
    16 & 76.4 & 74.4 & 86.8 & 94.0 & 68.8 & 74.8 & 81.2 & 63.2 & 77.5 \\
    36 & 75.5 & 70.8 & 84.4 & 91.2 & 64.8 & 71.2 & 77.6 & 66.0 & 76.2 \\
    \midrule
    \rowcolor{gray!10}\textbf{Number of frames} & & & & & & & & & \\
    2 & 72.4 & 70.8 & 86.0 & 92.8 & 67.2 & 70.0 & 67.2 & 50.4 & 72.1 \\
    4 & 74.4 & 72.0 & 87.2 & 92.4 & 66.8 & 66.0 & 72.8 & 58.4 & 73.8 \\
    8 & 76.4 & 74.4 & 86.8 & 94.0 & 68.8 & 74.8 & 81.2 & 63.2 & 77.5 \\
    

\bottomrule
\end{tabular}%
}
\label{tab:ablation_result}
\end{table*}

\section{Reproducibility Checklist}

\qt{We answer the questions outlined in the AAAI reproducibility checklist, available at \url{https://aaai.org/aaai-conference/reproducibility-checklist/}, as follows:}

\begin{itemize}
    \item Includes a conceptual outline and/or pseudocode description of AI methods introduced (yes/partial/no/NA) \\
    \textbf{Yes}
    \item Clearly delineates statements that are opinions, hypothesis, and speculation from objective facts and results (yes/no) \\
    \textbf{Yes}
    \item Provides well marked pedagogical references for less-familiare readers to gain background necessary to replicate the paper (yes/no)
    \\
    \textbf{Yes}
    \item Does this paper make theoretical contributions? (yes/no)
    \\
    \textbf{No}
    \item Does this paper rely on one or more datasets? (yes/no)
    \\
    \textbf{Yes}
    \item A motivation is given for why the experiments are conducted on the selected datasets (yes/partial/no/NA)
    \\
    \textbf{Yes}
    \item All novel datasets introduced in this paper are included in a data appendix. (yes/partial/no/NA)
    \\
    \textbf{Yes}
    \item All novel datasets introduced in this paper will be made publicly available upon publication of the paper with a license that allows free usage for research purposes. (yes/partial/no/NA)
    \\
    \textbf{Yes}
    \item All datasets drawn from the existing literature (potentially including authors’ own previously published work) are accompanied by appropriate citations. (yes/no/NA)
    \\
    \textbf{Yes}
    \item All datasets drawn from the existing literature (potentially including authors’ own previously published work) are publicly available. (yes/partial/no/NA)
    \\
    \textbf{Yes}
    \item All datasets that are not publicly available are described in detail, with explanation why publicly available alternatives are not scientifically satisficing. (yes/partial/no/NA)
    \\
    \textbf{Yes}
    \item Does this paper include computational experiments? (yes/no)
    \\
    \textbf{Yes}
    \item Any code required for pre-processing data is included in the appendix. (yes/partial/no).
    \\
    \textbf{Partial}. \kt{Only the explanation and diagram. To be made publicly available upon publication.} 
    \item All source code required for conducting and analyzing the experiments is included in a code appendix. (yes/partial/no)
    \\
    \textbf{Partial}. \kt{Only the explanation and diagram. To be made publicly available upon publication.}
    \item All source code required for conducting and analyzing the experiments will be made publicly available upon publication of the paper with a license that allows free usage for research purposes. (yes/partial/no)
    \\
    \textbf{Yes}
    \item All source code implementing new methods have comments detailing the implementation, with references to the paper where each step comes from (yes/partial/no)
    \\
    \textbf{Yes}
    \item If an algorithm depends on randomness, then the method used for setting seeds is described in a way sufficient to allow replication of results. (yes/partial/no/NA)
    \\
    \textbf{Yes}
    \item This paper specifies the computing infrastructure used for running experiments (hardware and software), including GPU/CPU models; amount of memory; operating system; names and versions of relevant software libraries and frameworks. (yes/partial/no)
    \\
    \textbf{Yes}
    \item This paper formally describes evaluation metrics used and explains the motivation for choosing these metrics. (yes/partial/no)
    \\
    \textbf{Yes}
    \item This paper states the number of algorithm runs used to compute each reported result. (yes/no)
    \\
    \textbf{Yes}
    \item Analysis of experiments goes beyond single-dimensional summaries of performance (e.g., average; median) to include measures of variation, confidence, or other distributional information. (yes/no)
    \\
    \textbf{Yes}
    \item The significance of any improvement or decrease in performance is judged using appropriate statistical tests (e.g., Wilcoxon signed-rank). (yes/partial/no)
    \\
    \textbf{Yes}
    \item This paper lists all final (hyper-)parameters used for each model/algorithm in the paper’s experiments. (yes/partial/no/NA)
    \\
    \qt{\textbf{Yes}. 
    }
    \item This paper states the number and range of values tried per (hyper-) parameter during development of the paper, along with the criterion used for selecting the final parameter setting. (yes/partial/no/NA)
    \\
    \textbf{Yes}. 
\end{itemize}
